%% file: neurips_2026.tex
\documentclass{article}

\usepackage[preprint]{neurips_2026}

\usepackage{xcolor}
\usepackage{colortbl}

\usepackage[most]{tcolorbox}
\usepackage{xcolor}

\definecolor{promptbg}{HTML}{F7F8FA}
\definecolor{promptborder}{HTML}{D9DDE3}
\definecolor{promptaccent}{HTML}{4F46E5}
\definecolor{prompttitlebg}{HTML}{EEF2FF}
\definecolor{prompttitlefg}{HTML}{312E81}

\newtcolorbox{llmprompt}[1][]{
  enhanced,
  breakable,
  colback=promptbg,
  colframe=promptborder,
  boxrule=0.4pt,
  arc=2mm,
  left=4mm,
  right=4mm,
  top=3mm,
  bottom=3mm,
  borderline west={1.5pt}{0pt}{promptaccent},
  fontupper=\sffamily,
  coltitle=prompttitlefg,
  attach boxed title to top left={xshift=3mm,yshift*=-\tcboxedtitleheight/2},
  boxed title style={
    colback=prompttitlebg,
    colframe=prompttitlebg,
    boxrule=0pt,
    arc=1mm,
    left=2mm,
    right=2mm,
    top=1mm,
    bottom=1mm
  },
  fonttitle=\sffamily\bfseries\footnotesize,
  drop fuzzy shadow=black!12,
  #1
}

\usepackage{xspace}

\usepackage{graphicx}

\usepackage{eso-pic}

\newcommand{\name}{\textsc{Peek}\xspace}

\usepackage[utf8]{inputenc} 
\usepackage[T1]{fontenc}    
\usepackage{hyperref}       
\usepackage{url}            
\usepackage{booktabs}       
\usepackage{amsfonts}       
\usepackage{nicefrac}       
\usepackage{microtype}      
\usepackage{xcolor}         

\usepackage{hyperref}
\usepackage{fontawesome5}
\usepackage{amsmath} 
\usepackage[ruled,vlined]{algorithm2e}
\usepackage{pifont}      
\usepackage[table]{xcolor}
\usepackage{multirow}
\usepackage{pifont}      
\usepackage{array}
\usepackage{makecell}
\usepackage{wrapfig}
\usepackage{caption} 

\usepackage{xurl}      
\usepackage{hyperref}  

\hypersetup{
}

\title{\name: Context Map as an Orientation Cache for Long-Context LLM Agents}

\author{
Zhuohan Gu$^{1}$,
\hspace{0.3em}
Qizheng Zhang$^2$,
Omar Khattab$^{1}$,
Samuel Madden$^{1}$\\
[0.3em]
$^1$MIT CSAIL \hspace{.3in} $^2$Stanford University\\
[0.4em]
{\ttfamily
\{{zgu15}, {okhattab}\}@mit.edu
\quad
{madden@csail.mit.edu}
}\\
[0.1em]
\href{https://github.com/zhuohangu/peek}{\faGithub\ \texttt{zhuohangu/peek}}
}

\begin{document}
\maketitle


\input{sections/abstract}

\input{macros}
\input{sections/intro}
\input{sections/background_motivation}
\input{sections/design}
\input{sections/eval}
\input{sections/related_work}
\input{sections/end}
\input{sections/acknowledgment}

\bibliographystyle{plain}
\bibliography{citations}

\appendix
\input{sections/appendix}







\end{document}

%% file: sections/abstract.tex
\begin{abstract}
Large language model (LLM) agents increasingly operate over long and recurring external contexts, like document corpora and code repositories. Across invocations, existing approaches preserve either the agent's trajectory, passive access to raw material, or task-level strategies. None of them preserves what we argue is most needed for repeated same-context workloads: reusable orientation knowledge
(e.g., what the context contains, how it is organized, and which entities, constants, and schemas have historically been useful) about the recurring context itself. We introduce \textbf{\name}, a system that caches and maintains this orientation knowledge as a context map: a small, constant-sized artifact in the agent's prompt that gives it a persistent peek into the external context. The map is maintained by a programmable cache policy with three modules: a Distiller that extracts transferable knowledge from inference-time signals, a Cartographer that translates it into structured edits, and a priority-based Evictor that enforces a fixed token budget. 
On long-context reasoning and information aggregation, \name\ improves over strong baselines by 6.3--34.0\% while using 93--145 fewer iterations and incurring 1.7--5.8$\times$ lower cost than the state-of-the-art prompt-learning framework, ACE. On context learning, \name\ improves solving rate and rubric accuracy by 6.0--14.0\% and 7.8--12.1\%, respectively, at 1.4$\times$ lower cost than ACE. These gains generalize across LMs and agent architectures, including OpenAI Codex, a production-grade coding agent. Together, these results show that a context map helps long-context LLM agents interact with recurring external contexts more accurately and efficiently.
\end{abstract}

%% file: macros.tex
\newcommand{\fillme}{{\bf XXX}\xspace}
\newcommand{\eg}{{\it e.g.,}\xspace}
\newcommand{\ie}{{\it i.e.,}\xspace}
\newcommand{\etal}{{\it et.~al}\xspace}
\newcommand{\bigO}{\mathrm{O}}

\newcommand{\cmark}{\ding{51}}  
\newcommand{\xmark}{\ding{55}}  

\definecolor{checkmark}{HTML}{00D100}
\definecolor{crossmark}{HTML}{FF5733}

\newcommand{\qz}[1]{{\color{purple}{(Qizheng: #1)}}}

\newenvironment{packeditemize}{\begin{list}{$\bullet$}{\setlength{\itemsep}{0.5pt}\addtolength{\labelwidth}{-4pt}\setlength{\leftmargin}{2ex}\setlength{\listparindent}{\parindent}\setlength{\parsep}{1pt}\setlength{\topsep}{2pt}}}{\end{list}}
\newcounter{packednmbr}
\newenvironment{packedenumerate}{\begin{list}{\thepackednmbr.}{\usecounter{packednmbr}\setlength{\itemsep}{0.5pt}\addtolength{\labelwidth}{-4pt}\setlength{\leftmargin}{2ex}\setlength{\listparindent}{\parindent}\setlength{\parsep}{1pt}\setlength{\topsep}{2pt}}}{\end{list}}

\newcommand{\todo}[1]{\textcolor{red}{#1}}

\definecolor{lightnavyblue}{rgb}{0.80, 0.90, 1.00}
\newtcolorbox{insightbox}{
    sharpish corners, 
    colback=lightnavyblue!35, 
    colframe=black, 
    boxrule=0.8pt, 
    toprule=2pt, 
    bottomrule=2pt, 
    enhanced,
    drop fuzzy shadow, 
    fuzzy shadow={0.5mm}{-0.5mm}{-0.2mm}{0.3mm}{black!35} 
}

\definecolor{boxbackground}{rgb}{0.95, 0.95, 0.95} 
\definecolor{boxborder}{rgb}{0.3, 0.3, 0.3} 

%% file: sections/intro.tex
\section{Introduction}
\label{sec:intro}

\begin{figure}[!ht]
    \centering
    \includegraphics[width=\linewidth]{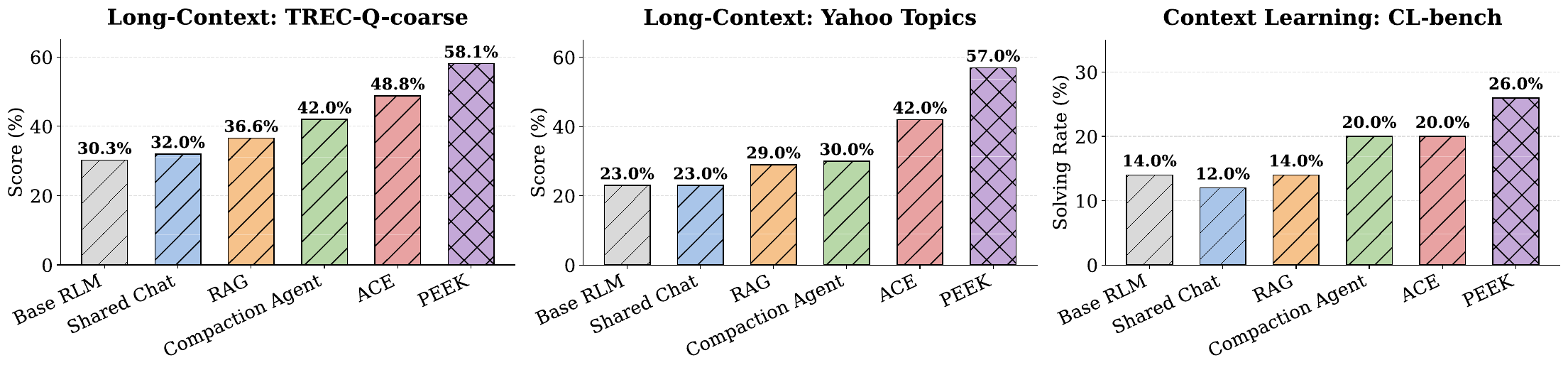}
    \caption{\textbf{Performance Snapshot (GPT-5-mini as the Base LM).} \name (our system) consistently achieves the highest scores across long-context tasks compared with strong baselines.}
    \label{fig:perf_snapshot}
\end{figure}

Large language model (LLM) agents such as Claude Code~\cite{anthropic2025claudecode}, Codex~\cite{openai_codex_cli}, RLM~\cite{zhang2025recursive}, OpenClaw~\cite{openclaw2025}, and Hermes Agent~\cite{nousresearch2026hermes} increasingly operate over large and recurring \emph{external contexts}: document corpora, code repositories, or other resources that the agent repeatedly queries but live outside the LLM's context window. 
For example, an enterprise data analyst might use LLM agents to repeatedly query a corpus of 50k+ user feedback entries with questions such as \emph{``Do users prefer feature A or B?''} or \emph{``What onboarding complaints appear most often?''}. 
In these workloads, the external context stays largely the same while the tasks change.
Modern agentic systems manage such contexts in several ways: holding it directly in a long context window~\cite{yang2025context, yao2025training, lee2502infinitehip, kang2025acon, yu2025memagent, wu2025resum}, retrieving relevant pieces on demand (Retrieval-Augmented Generation (RAG)~\cite{lewis2020retrieval, ray2025metis}), offloading it into an external environment that the agent inspects, slices, decomposes, and queries at inference time~\cite{zhang2025recursive, anthropic2025claudecode, openclaw2025}, or compacting the external context so it can fit into the context window~\cite{yu2025memagent, wu2025resum, sun2510scaling}.
Orthogonally, \emph{prompt learning} (also called \emph{context engineering})~\cite{zhang2510agentic, agrawal2025gepa, suzgun2025dynamic, shinn2024reflexion} treats the prompt as a learnable object, accumulating task-level rules, reflections, or procedures without updating model weights.

These methods are necessary, but each preserves a different kind of object, and none preserves what we argue is most needed for repeated same-context workloads. 
Shared chat carries the agent's prior trajectory, which can support in-context learning but rapidly accumulates into noisy, low-density context. 
History compaction summarizes chat history to keep an agent aware of prior activity, but discards reusable knowledge about the external context. 
RAG, context offloading, and context compaction preserve passive access to raw or condensed material, but do not maintain a curated artifact about the context itself. 
Prompt learning curates task-level strategies, not contextual knowledge about a recurring external context. 
We argue that what is missing is a bounded, managed artifact that preserves reusable \emph{orientation knowledge} about a recurringly queried external context.

We propose to fill this gap with a \emph{context map}: a small, constant-sized artifact inside the agent's prompt that stores reusable orientation knowledge about the external context. Orientation knowledge is the reusable map a human analyst would keep after several passes through a corpus: what the corpus contains, how it is organized, which constants or schemas have historically proved most useful, and where different kinds of evidence live.
It is not simply a summary of the current conversation or a task strategy. 
Instead, inspired by hardware caches and database indexes, a context map is a small, prompt-resident view of a larger external context.
Unlike KV-cache management~\cite{zandieh2025turboquant, li2025commvqcommutativevectorquantization, du2026bitdecoding, liu2024droidspeak}, which is a model-level optimization that reuses or compresses token states, a context map is an agent-level artifact whose value comes from curated meaning. 
The map is never compacted away or externalized into environment storage; it persists across queries and is updated as the agent interacts with the external context.

This paper tackles two concrete questions: given a constant-sized context map that sits in the system message, \emph{how should we decide what goes in it} and \emph{how much does it help the agentic system make sense of the larger external context more efficiently and reliably?}
We present \textbf{\name}, a system that maintains and evolves such a context map: our map is a fixed-budget \emph{peek} into the external context that helps the agent interpret, navigate, and reason about it.
\name runs the underlying agent on user tasks and updates the map from inference-time signals through three modules: a \emph{Distiller} that extracts transferable contextual knowledge from execution trajectories, a \emph{Cartographer} that translates these into structured edits, and an \emph{Evictor} that enforces a fixed token budget via priority-based eviction. Separating extraction from editing is essential: without it, non-contextual facts leak into the map and updates become noisy and duplicative (\S\ref{subsec:ablations}).

We evaluate \name on two categories of long-context tasks that most stress repeated context interaction: (1) reasoning and aggregation~\cite{bertsch2025oolong}, which requires identifying and combining evidence distributed across the external context, and (2) context learning~\cite{dou2026cl}, which requires acquiring task-relevant knowledge from the external context and applying it across related tasks.
Our key findings:
\begin{packeditemize}
    \item \textbf{Consistent Quality Gains.} Across all baselines, \name improves OOLONG by 6.3--34.0\% and CL-bench by 6.0--14.0\% / 7.8--12.1\% in solving rate / rubric accuracy. Compared to the SOTA prompt-learning framework, \name gains 10.7\% on OOLONG and 6.0\% / 9.9\% on CL-bench solving rate / rubric accuracy. A performance snapshot is shown in Figure~\ref{fig:perf_snapshot}.
    \item \textbf{Fewer Iterations.} \name lies on the iteration--quality Pareto frontier across all benchmarks. On OOLONG, \name uses 93--145 fewer iterations than the SOTA prompt-learning framework.
    \item \textbf{Better Cost--Quality Tradeoff.} \name lies on the cost--quality Pareto frontier across all benchmarks, costing 1.7--5.8$\times$ less than SOTA prompt learning on OOLONG and 1.4$\times$ less on CL-bench.
    \item \textbf{Strong Generalization.} \name generalizes across LMs and agents: gains hold with the latest frontier model (GPT-5.5) and an open-source LM (Qwen3-Coder-Next-FP8), and they also transfer to a different backbone agent (OpenAI Codex).
\end{packeditemize}

%% file: sections/background_motivation.tex
\section{Background and Motivation}
\label{sec:background}
\subsection{The Design Space of Context Management}
\begin{figure}[!h]
    \centering
    \includegraphics[width=\linewidth]{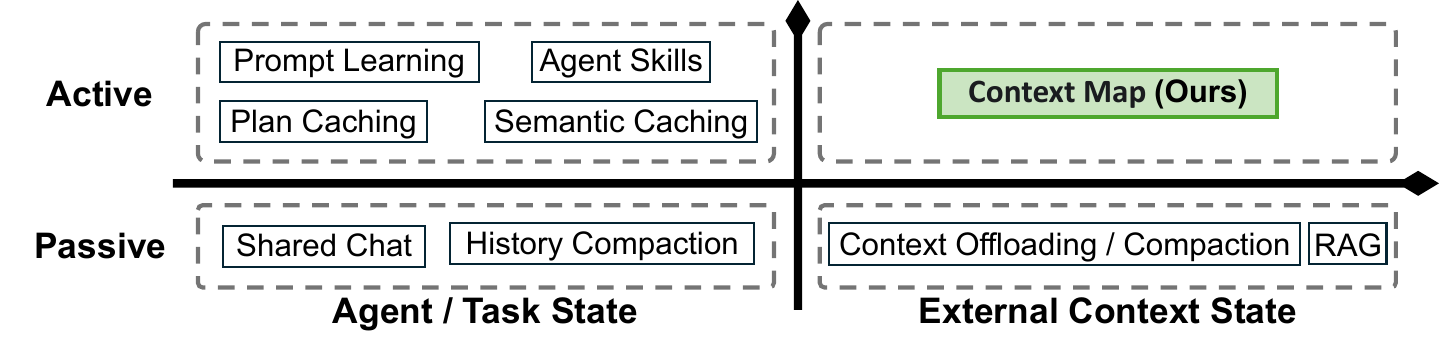}
    \caption{\textbf{Design Space of Agent State.} Context maps fill the active external-context quadrant.}
    \label{fig:quadrants}
\end{figure}

Figure~\ref{fig:quadrants} shows related context-management methods along two axes.  The horizontal axis captures whether the method is about managing \emph{Agent / Task State}, i.e., the agent's execution or task behavior, or about managing \emph{External Context State}, i.e., the recurring external context itself. On the vertical axis, \emph{Active} methods deliberately maintain an artifact across interactions while \emph{Passive} methods carry, retrieve, or summarize state when the transcript, query, or context-window pressure requires it.

Representative active agent/task-state methods include prompt learning~\cite{zhang2510agentic, agrawal2025gepa, shinn2024reflexion, yuksekgonul2025optimizing}, which improves LM task performance without updating model weights by treating the prompt as a learnable object, agent skills~\cite{agentskills_overview, anthropic_agent_skills,  openai_codex_agent_skills, cursor_agent_skills}, agentic plan caching~\cite{zhang2026agenticplancachingtesttime}, and semantic caching~\cite{bang-2023-gptcache, schroeder2026vcacheverifiedsemanticprompt, amazon-semantic-cache}, which maintain task-facing procedures, plans, or answers.
In passive agent/task state, shared chat carries chat history, and history compaction summarizes the execution record when the context window fills to keep the current task alive.
Passive external-context methods include RAG~\cite{lewis2020retrieval, ray2025metis}, which retrieves query-relevant chunks from the external context, context offloading, which lets the LM inspect and query externalized material at inference time, and context-compaction methods~\cite{yu2025memagent, sun2510scaling, wu2025resum} such as MemAgent~\cite{yu2025memagent}, which condense external context so it can be brought into the context window.
The missing quadrant is active external-context state: i.e., methods that actively maintain an  artifact that captures what has been learned about a recurring external context.  This is the gap that \name fills.

\subsection{Gap: Context Maps for Orientation Knowledge}
\label{subsec:context_maps_gap}

Returning to the example in \S\ref{sec:intro}, consider an enterprise analyst who uses LLM agents to repeatedly query the same 50k-entry feedback corpus. Reflecting on what a human might do with such a collection of documents after a few previous rounds of analysis, they would not start each question from scratch. They might keep a lightweight table of contents, memos about key entities and constants, a record of which regions have been inspected, and answers to common intermediate results.
An agent facing the same setting needs an analogous aid, which we posit can be a small maintained view of the external context that stays resident in the prompt. This view should seek to help the agent understand what the external context contains, how it is organized, and such additional notes. It should also remain bounded, to create a caching advantage relative to accessing the corpus, which might in contrast be arbitrarily large. 
The rest of this paper studies two key questions inside the active external-context quadrant: what belongs in a small prompt-resident context map, and how much does it help agentic systems when answering sequences of requests over a shared large corpus?

%% file: sections/design.tex
\section{Context Map as a Cache}
\label{sec:design}
\begin{figure}[!t]
    \centering
\includegraphics[width=\linewidth]{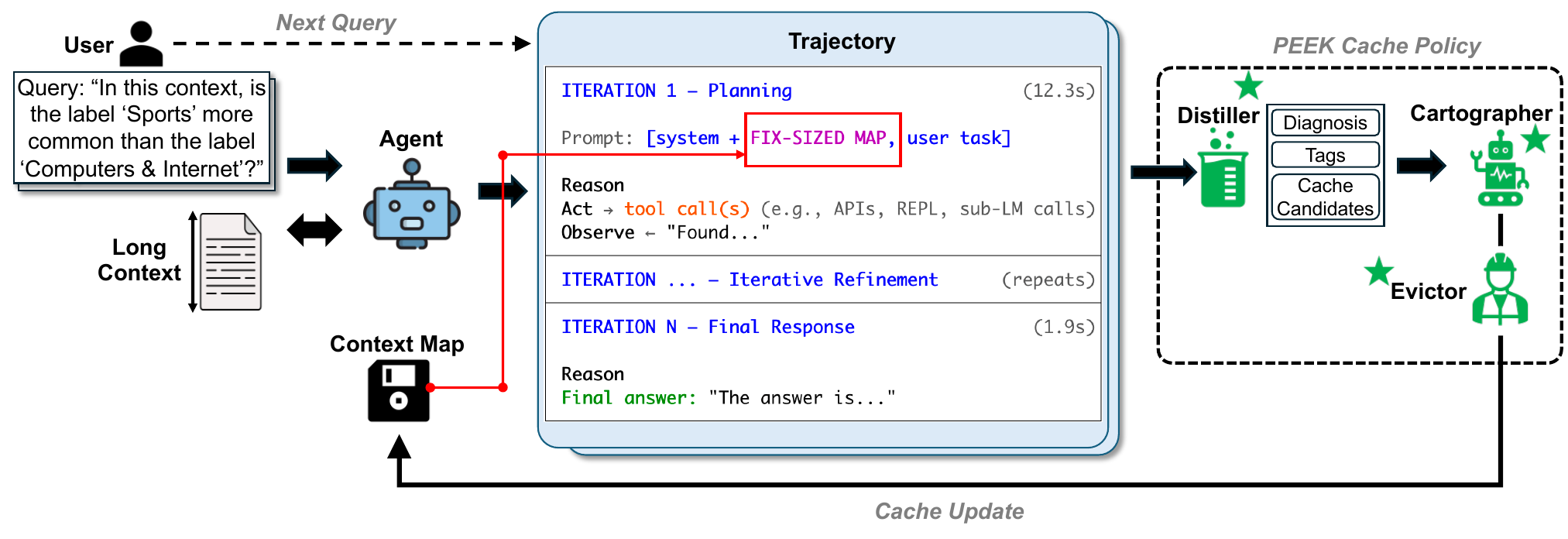}
    \caption{\textbf{The \name\ System.} Inspired by caching in computer systems and the notion of peeking, \name\ caches orientation knowledge in a context map and updates it through a modular process consisting of a Distiller, a Cartographer, and an Evictor.}
    \label{fig:design}
\end{figure}

We present \name, our system for building and maintaining the context map. The map resides inside the LM's prompt and contains transferable knowledge and understanding about an external context with which the agent repeatedly interacts. We treat the context map as a \emph{cache} inspired by the idea of a cache in computer systems, where fast memory near the processor keeps frequently used data close at hand. Here, the cache is a small piece of information supplied in addition to the larger external context, allowing the LM to make sense of the external context more efficiently and reliably.

As shown in Figure~\ref{fig:design}, \name implements this cache through two coupled mechanisms. The first is a constant-sized \emph{context map} (boxed in red) that sits in the agent's system prompt. The second is a \emph{cache management policy} (dashed box, green stars) that, after each query completes, inspects the execution trajectory and updates the map for use in the next query. The cache evolution can be frozen after as few as $m{=}1$ query (e.g., once sufficient knowledge about the external context has been cached for subsequent queries), after which the map is simply reused without further updates.

The design addresses three questions in turn. First, \emph{what should go in the map~(\S\ref{subsec:peeking})?} The context map provides a compact ``peek'' into the external context. Second, \emph{how does the map evolve~(\S\ref{subsec:evolve})?} A \emph{Distiller} diagnoses what the agent learned, then a \emph{Cartographer} applies structured edits. Separating the two is essential: without a Distiller, task-specific facts leak into the cache, while without a Cartographer, updates become noisier, duplicative, and prone to overwriting stable entries. Third, \emph{how is the map kept compact~(\S\ref{subsec:evictor})?} A priority-based \emph{Evictor} enforces a fixed token budget. We ablate our design choices in \S\ref{subsec:ablations}.

\subsection{Context Map: Peeking into Context}
\label{subsec:peeking}
\begin{figure}[!ht]
    \centering
\includegraphics[width=\linewidth]{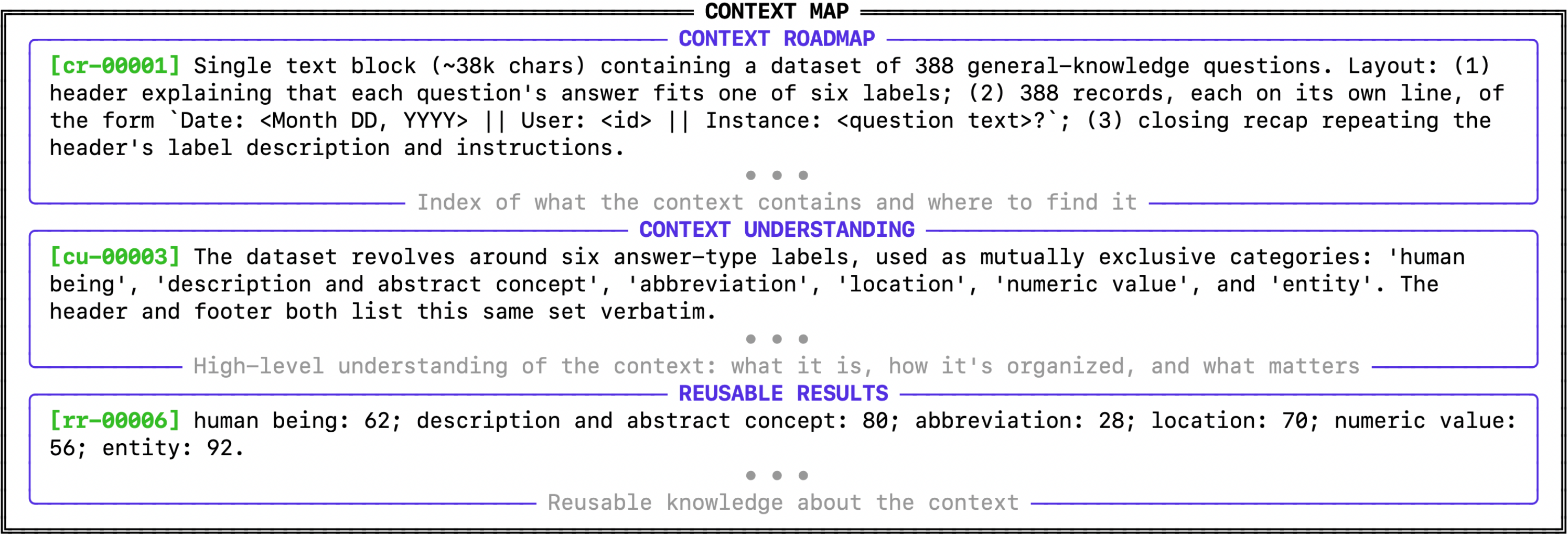}
    \caption{\textbf{Example Context Map Generated by \name\ (Partially Shown).} The map stores contextual knowledge in structured sections with stable item IDs, enabling consistent cache updates.}
    \label{fig:context map example}
\end{figure}
When an agent repeatedly interacts with a long external context, it often spends its first several iterations building a working understanding of that context: what it contains, how it is organized, which entities and concepts matter, and how to navigate it. This orientation work is transferable: it helps answer any query about the same external context. Interaction history is not a substitute: it is a long, noisy execution log, so reusable orientation knowledge is sparse and easily summarized or discarded when the window fills. Persisting the full history actually hurts performance (\S\ref{subsec:main_results}).

The context map preserves this orientation knowledge. Concretely, our default context map has two sections: the \texttt{Context Roadmap}, an abbreviated index of what the external context contains and where, and the \texttt{Context Understanding} which consists of a higher-level description of the context including key entities, concepts, and their relationships.
Three optional sections are populated as needed: \texttt{Domain Constants} for exact values or enumerated sets, \texttt{Reusable Results} for derived computations, and \texttt{Parsing Schema} for format and delimiter structure. All five sections start nearly empty (i.e., just headers; shown in Appendix~\ref{sec:initial_map}) and the map can even begin as a blank slate. Figure~\ref{fig:context map example} shows an example of the map after one query, making the contrast with the initial map explicit. Here, for example, a \texttt{Context Roadmap} entry caches a navigational summary ``\texttt{[cr-00001]} Single text block ($\sim$38k chars) containing 388 records of the form\ldots,'' which any future query on the same corpus can reuse without re-discovering. We deliberately avoid pre-populating or hand-crafting entries, as doing so would undercut the core premise of \name: it should discover and accumulate contextual knowledge automatically through interaction.
\subsection{Map Evolution: Programmable Cache Policy}
\label{subsec:evolve}

\definecolor{hlblue}{RGB}{205,232,241}
\definecolor{hlorange}{RGB}{244,223,205}
\definecolor{hlgreen}{RGB}{220,239,220}
\definecolor{hlpurple}{RGB}{232,220,245}

\setlength{\columnsep}{0.8em}
\begin{wrapfigure}[9]{r}[0.057\textwidth]{0.56\textwidth} 
\vspace{-13pt}
  \begingroup
  \small
  \setlength{\fboxsep}{0.1pt}

  \setlength{\algomargin}{0.15em}

  \SetInd{0.25em}{0.55em}

  \SetAlgoHangIndent{0.5em}

  \SetAlgoSkip{}
  \SetAlgoInsideSkip{}

  \SetCustomAlgoRuledWidth{0.5\textwidth}
  \newcommand{\hlfn}[1]{\colorbox{hlpurple}{$\texttt{#1}$}}
  \newcommand{\smap}{\text{sys}{+}\boldsymbol{\textit{map}}}

  \SetAlgoVlined
  \SetVlineSkip{-0.02em}

  \begin{algorithm}[H]
    \caption{\name Programmable Cache Policy}
    \label{alg:system}
    \KwIn{context $\mathcal{C}$, $Q_{1:n}$, budget $B$, evolve steps $m\!\le\! n$}
    $\textit{map} \leftarrow \texttt{Init}()$\;
    \For{$i \leftarrow 1$ \KwTo $n$}{
      $(\textit{a}_i,\textit{traj}) \leftarrow \texttt{AgentLoop}([\smap,Q_i],\mathcal{C})$\;
      \If{$i \le m$}{
        $(\textit{diag},\textit{tags},\textit{cands}) \leftarrow
          \hlfn{Distiller}(\textit{traj},\textit{map})$\;
        $\textit{edits} \leftarrow
          \hlfn{Cartographer}(\textit{diag},\textit{tags},\textit{cands};\textit{map})$\;
        $\textit{map} \leftarrow \texttt{Apply}(\textit{map},\textit{edits})$\;
        $\textit{map} \leftarrow
          \hlfn{Evictor}(\textit{map}, B)$\;
      }
    }
  \end{algorithm}
  \endgroup
  \vspace{-10pt}
\end{wrapfigure}

Algorithm~\ref{alg:system} runs an agent over $n$ tasks on the same recurring external context $\mathcal{C}$. The current map is prepended to every agent run, but updates are performed only for the first $m$ runs. Thus $m{=}n$ gives fully online adaptation.

\paragraph{Distiller} Each agent run produces an 
\label{subsec:distiller}
\emph{execution trajectory}: the reasoning steps, actions (e.g., tool calls, API calls, or sub-agent calls), and observations the agent emits, whose exact structure varies with the underlying model and agent architecture (Figure~\ref{fig:design}). The \emph{Distiller} takes this trajectory and the current context map and produces three outputs: a \emph{diagnosis} of how the agent spent its iterations (orientation versus task-specific work, and where it stalled or succeeded); per-item \emph{tags} for current context map entries (\texttt{helpful}, \texttt{harmful}, \texttt{neutral}, or \texttt{stale} with respect to this run); and \emph{cache candidates}: only transferable contextual information built up during execution is retained, and task-specific rules are discarded. By default, the Distiller operates without ground truth or a final answer, relying solely on execution signals that are always available.
The rationale is that an execution trajectory contains rich information about the agent's interaction with the context, but it is long and noisy---prohibitive, often impossible, for a human to review manually. Modern LMs, however, are very good at trajectory analysis~\cite{zhuge2024agent, ou2025agentdiagnose, lu2025agentrewardbench}. We found this trajectory-driven diagnosis substantially more effective than alternatives we explored, such as runtime feedback and behavioral instructions. We discuss approaches we tried that did not work in detail in Appendix~\ref{subsec:Things We Tried that Did Not Work}.

\paragraph{Cartographer.} To update the map, the \emph{Cartographer} translates the Distiller's output into structured edit operations (\textsc{Add}, \textsc{Delete}, or \textsc{Replace}) against the current context map. Each map item carries a unique identifier, keeping updates local and traceable. The Cartographer deduplicates incoming candidates against existing entries and emits a minimal edit set that increases overall cache value.
\label{subsec:cartographer}

\paragraph{Evictor: Priority-based Eviction.} The context map has a hard token budget $B$, fixed across all queries on a context and across all experiments. The \emph{Evictor} enforces this budget after each cache update: if the map exceeds $B$, it evicts items in ascending order of their Distiller-accumulated scores, breaking ties by removing older entries first. Eviction follows the inverse of the section-value hierarchy from \S\ref{subsec:peeking}: \texttt{Parsing Schema} entries go first, then \texttt{Reusable Results}, then \texttt{Domain Constants}, while \texttt{Context Roadmap} and \texttt{Context Understanding} are protected until last.
\label{subsec:evictor}

%% file: sections/eval.tex
\section{Experiments}
\label{sec:eval}
In this section, we present our experimental evaluation of \name.
Our key findings are:
\begin{packeditemize}
    \item \textbf{Consistent Quality Gains (\S\ref{subsec:main_results}).} \name outperforms strong baselines on every benchmark and metric, with gains of 6.3--34.0\% on OOLONG and 6.0--14.0\% / 7.8--12.1\% in solving rate / rubric accuracy on CL-bench.
    \item \textbf{Fewer Iterations (\S\ref{subsec:main_results}).} The context map helps the agent make sense of the context more efficiently and better, consistently occupying the high-quality, low-iteration region across benchmarks. 
    \item \textbf{Better Cost--Quality Tradeoff (\S\ref{subsec:main_results}).} \name lies on the cost--quality Pareto frontier across all benchmarks, delivering substantially better performance per dollar than other baselines. 
    \item \textbf{Strong Generalization Across Models and Agents (\S\ref{subsec:sensitivity}).} Gains are consistent when replacing the base LM (GPT-5-mini $\rightarrow$ GPT-5.5 and Qwen3-Coder-Next-FP8) and when swapping the backbone agent (RLM $\rightarrow$ Codex), confirming that \name is not tied to a specific model or agent.
\end{packeditemize}

\subsection{Datasets and Metrics}
\label{subsec:datasets}
We evaluate \name on tasks where we can simulate multiple long contexts, each containing multiple tasks that query the same context. We use benchmarks that are natively structured this way and require no data engineering or self-construction. Our evaluation covers two categories of long-context demands: (1) reasoning and information aggregation over long inputs~\cite{bertsch2025oolong}, which requires identifying and combining distributed evidence, and (2) context learning~\cite{dou2026cl}, which requires acquiring new task-relevant knowledge from the context and applying it across tasks.

\paragraph{OOLONG \cite{bertsch2025oolong}.} A long-context reasoning benchmark. Each task requires the model to identify relevant segments in the long context and aggregate them. We follow the evaluation from the original paper: $\mathrm{score}(\hat{y}) = 0.75^{|y-\hat{y}|}$ for numerical answers and exact match otherwise. Of OOLONG's 10 categories, we focus on the three hardest splits: \texttt{trec\_coarse}, \texttt{agnews}, and \texttt{yahoo}.

\paragraph{CL-bench~\cite{dou2026cl}.} A recent, highly challenging context-learning benchmark spanning domain-specific knowledge, rule systems, complex procedures, and laws. Each context is paired with multiple related tasks (up to 12). We follow the original all-or-nothing evaluation setup, using GPT-5.1 as the judge model, and report both solving rate (coarse) and rubric accuracy (fine-grained).

\subsection{Baselines and Methods}
\label{subsec:baselines}
We implement \name on top of the official RLM~\cite{zhang2025recursive} system, which stores contexts as environment variables and naturally provides an externalized-context interface. All baselines and methods are built on RLM to ensure a fair comparison. Because applying the agent to millions of tokens with the latest frontier model is extremely costly (as shown in \S\ref{subsec:sensitivity}), we use GPT-5-mini~\cite{singh2025openai} as the base LM in our main results (\S\ref{subsec:main_results}) and perform some less exhaustive evaluations with GPT-5.5 and other open-source models. Specifically, in our generalizability and sensitivity analysis (\S\ref{subsec:sensitivity}), we evaluate the latest frontier model (GPT-5.5~\cite{openai2026gpt55systemcard}), an open-source model (Qwen3-Coder-Next~\cite{cao2026qwen3}), and an alternative backbone agent (OpenAI Codex~\cite{openai_codex_cli}), observing consistent gains. Besides base RLM, we compare \name against common context-offloading methods and SOTA prompt-learning frameworks for selecting what to include in the map, as follows:

\paragraph{Recursive Language Models (RLMs)~\cite{zhang2025recursive}.}
\label{rlm}
RLM is a recent, widely used agent that externalizes contexts as variables in a Read-Eval-Print Loop (REPL) environment and has LMs write code to inspect, decompose, and recursively call themselves on selected snippets. We follow the official usage released by the authors and more details are provided in Appendix \ref{subsec:RLMs}.

\paragraph{RLMs w/ Shared Chat.} This method runs multiple questions within a single, continuing RLM conversation, instead of restarting a fresh \texttt{rlm.completion()} for every task. By stacking tasks in one chat window, later tasks can see the previous trajectory and leverage in-context learning.

\paragraph{Retrieval-Augmented Generation (RAG)~\cite{lewis2020retrieval}.} RAG is a method in which an LM answers a query by first retrieving relevant pieces of information (chunks) from an external corpus and then incorporating them into the prompt. In our setup, we use a standard RAG pipeline with a modern embedding model (text-embedding-3-small~\cite{openai_text_embedding_3_small}). Specifically, we split the context into chunks, embed and index them, and retrieve the top-$k$ chunks by query similarity at run time for the prompt.

\paragraph{Compaction Agent.} We use MemAgent~\cite{yu2025memagent} off the shelf to summarize contexts, following its prescribed chunk-based rolling-memory pipeline and prompts (Appendix \ref{sec:MemAgent}) without modification. Concretely, the context is processed sequentially in 5k-token windows, and a running memory is updated iteratively. Due to the extremely high cost of applying this process iteratively to millions of tokens, we use GPT-5.4-nano~\cite{openai_gpt54_nano} as the base LM for compaction.

\paragraph{Agentic Context Engineering (ACE)~\cite{zhang2510agentic}.} ACE is the current state-of-the-art framework for prompt learning, and we use it by following the pipeline and prompts released by the authors (Appendix \ref{sec:prompt-ace-reflector}, \ref{sec:prompt-ace-curator}) without modification. ACE optimizes LM contexts by treating them as evolving, structured playbooks. Through generation, reflection, and curation, it continuously accumulates, refines, and organizes task-specific strategies via incremental updates based on execution feedback.
\subsection{Main Results}
\label{subsec:main_results}

\definecolor{mygreen}{RGB}{0,128,0}
\definecolor{myred}{RGB}{200,40,40}
\newcommand{\gain}[1]{\textcolor{mygreen}{$_{#1}$}}
\newcommand{\lose}[1]{\textcolor{myred}{$_{#1}$}}

\begin{table*}[!t]
\centering
\renewcommand{\arraystretch}{1.12}
\setlength{\tabcolsep}{3pt}

\setlength{\heavyrulewidth}{2pt}
\setlength{\lightrulewidth}{0.8pt}
\setlength{\cmidrulewidth}{0.8pt}

\fontsize{9}{11}\selectfont

\begin{tabular}{p{4.8cm} ccccc}
\toprule
\multirow{2}{*}{\textbf{Method}} &
\multicolumn{3}{c}{\textbf{OOLONG}} &
\multicolumn{2}{c}{\textbf{CL-bench}$\uparrow$} \\
\cmidrule(lr){2-4} \cmidrule(lr){5-6}
& \textbf{TREC-Q-coarse$\uparrow$} & \textbf{AGNews$\uparrow$} & \textbf{Yahoo$\uparrow$}
& \textbf{Solve$\uparrow$} & \textbf{Rubric$\uparrow$} \\
\specialrule{\heavyrulewidth}{0pt}{0pt}

\rowcolor{gray!15}
\multicolumn{6}{c}{\texttt{GPT-5-mini-2025-08-07 as Base LM}} \\
\textcolor{gray}{RLM} & \textcolor{gray}{30.3} & \textcolor{gray}{46.5} & \textcolor{gray}{23.0} & \textcolor{gray}{14.0} & \textcolor{gray}{54.5} \\
\midrule

RLM + \textbf{Shared Chat} & 32.0\gain{+1.7} & 49.6\gain{+3.1} & 23.0\gain{+0.0} & 12.0\lose{-2.0} & 51.3\lose{-3.2} \\
RLM + \textbf{RAG} & 36.6\gain{+6.3} & 63.1\gain{+16.6} & 29.0\gain{+6.0} & 14.0\gain{+0.0} & 55.6\gain{+1.1} \\
RLM + \textbf{Compaction Agent} & 42.0\gain{+11.7} & 49.5\gain{+3.0} & 30.0\gain{+7.0} & 20.0\gain{+6.0} & 54.6\gain{+0.1} \\
RLM + \textbf{ACE} (\emph{Online Adaptation}) & 48.8\gain{+18.5} & 61.6\gain{+15.1} & 42.0\gain{+19.0} & 20.0\gain{+6.0} & 53.5\lose{-1.0} \\
RLM + \textbf{\name} & \textbf{58.1}\gain{+\textbf{27.8}} & \textbf{69.4}\gain{+\textbf{22.9}} & \textbf{57.0}\gain{+\textbf{34.0}} & \textbf{26.0}\gain{+\textbf{12.0}} & \textbf{63.4}\gain{+\textbf{8.9}} \\
\bottomrule
\end{tabular}
\caption{\textbf{Results of Different Methods Across Long-context Benchmarks.} All methods are built on top of RLM using GPT-5-mini as the base LM. Green and red subscripts denote absolute improvement and degradation over the base RLM, respectively. \name outperforms all baselines across every metric.}
\label{tab:main_results}
\end{table*}

As shown in Table~\ref{tab:main_results}, \name consistently outperforms all baselines across benchmarks. On OOLONG, \name beats the SOTA prompt-learning method ACE by $+7.8$--$15.0\%$. On CL-bench, \name improves over ACE by +6.0\% solving rate and +9.9\% rubric accuracy. These gains are consistent across both coarse metrics (solving rate) and fine-grained ones (rubric accuracy), demonstrating that the context map improves task correctness by strengthening context understanding rather than overfitting to specific task types. Importantly, all reported context-map variants suffice to show improvements with m $\leq 4$, while ACE is run in full online-adaptation mode (reflector+curator calls on every query), yet \name remains best in both quality and efficiency.

The comparison with individual baselines clarifies where existing methods break down. Shared Chat, which persists the full conversation history in the chat window, gives only small gains on OOLONG and even hurts CL-bench, confirming that accumulating raw context becomes noise that degrades rather than aids. RAG helps the agent glimpse the context when it is well structured, as in OOLONG, but falls short on less structured, more complex settings like CL-bench. Compaction Agent improves coarse success but barely improves CL-bench rubric accuracy. ACE shows a clear tradeoff: solving rate improves by $6.0\%$, but rubric accuracy drops by $1.0\%$, consistent with a playbook that indicates task-specific optimization that trades off partial correctness elsewhere. Together, these results show that \name is a powerful and reliable system for unleashing the full potential of agents in repeated interaction with recurring external contexts, without requiring labeled supervision.

\begin{figure}[!ht]
    \centering
    \includegraphics[width=\linewidth]{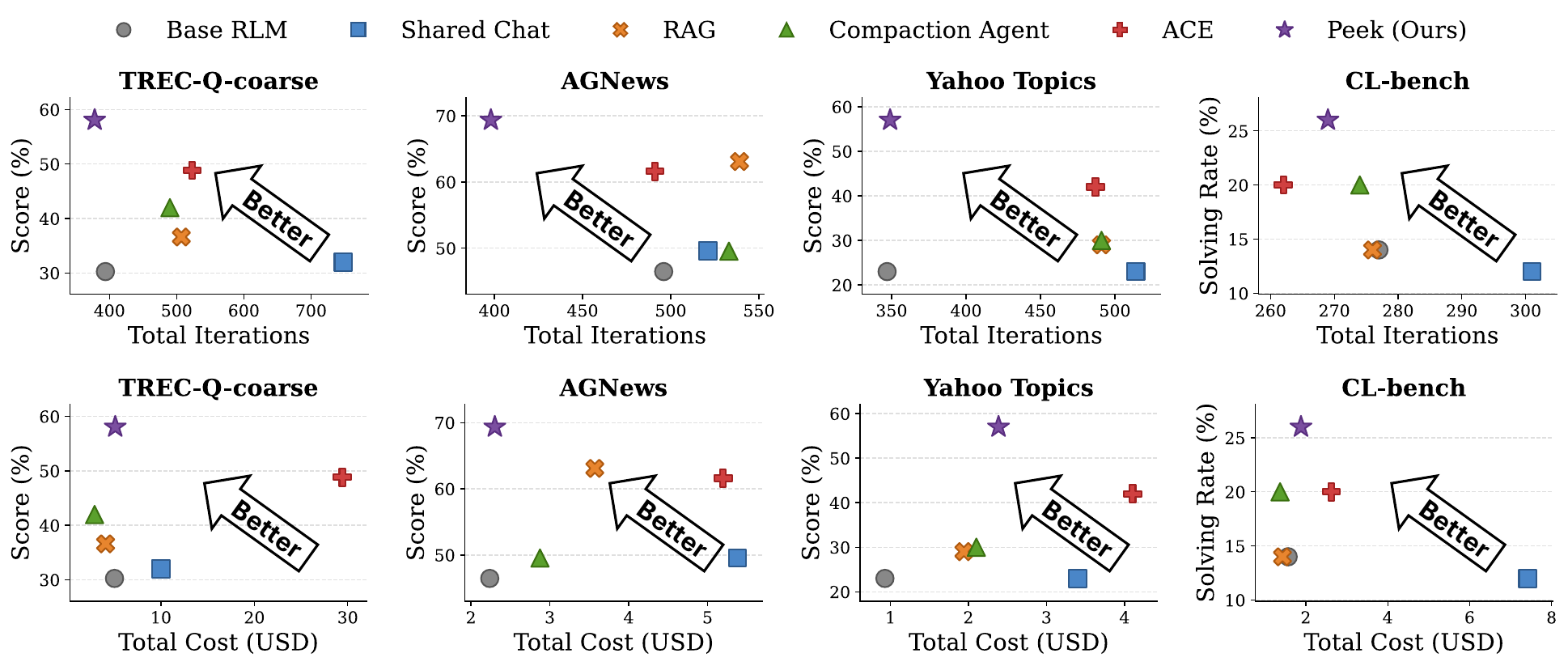}
    \caption{\textbf{Score vs.\ Total Iterations (Top):} The upper-left region (higher score, fewer iterations) is better. \textbf{Score vs.\ Total Cost (Bottom):} Total cost includes both execution cost and method-specific overhead (ACE adaptation or \name maintenance), and the upper-left region (higher score, lower cost) is better. Across both views, \name consistently lies on the Pareto frontier across all four benchmarks.}
\label{fig:score_vs_iteration_and_score}
\vspace{-.1in}
\end{figure}

Figure~\ref{fig:score_vs_iteration_and_score} (top) shows that the baselines make different runtime efficiency–quality trade-offs, but none match \name. Shared Chat drives iterations sharply upward (up to 748 on OOLONG and 301 on CL-bench) for negligible or negative quality gains. RAG and Compaction Agent use moderate iterations (490--539) with only modest quality improvements. ACE requires 93--145 more iterations than \name on OOLONG while scoring 7.8--15.0\% lower. On CL-bench, ACE uses the fewest iterations (262) but trails \name by $+6.0\%$ solving rate and $+9.9\%$ rubric accuracy, suggesting that its low iteration count reflects premature shortcuts rather than efficiency. In contrast, \name sits at or near the fewest iteration levels on all benchmarks while significantly leading in quality.

Besides inference-time efficiency, we also evaluate score versus total cost in Figure~\ref{fig:score_vs_iteration_and_score} (bottom), where total cost includes both execution cost and method-specific overhead. On OOLONG, \name stays near base-RLM cost on TREC-Q-coarse and AGNews while improving accuracy by $22.9$--$27.8\%$. On Yahoo Topics, \name has a mid-range cost among the methods while achieving the largest accuracy gain ($+34.0\%$). ACE costs $1.7$--$5.8\times$ more than \name yet scores lower on all three splits. Shared Chat costs $1.4$--$2.3\times$ more with far lower quality. RAG and Compaction Agent can be cheaper on some splits but trail \name by $6.3$--$28.0\%$. On CL-bench, Shared Chat costs $3.9\times$ more than \name while hurting quality, ACE costs $1.4\times$ more while trailing on both metrics, and RAG and Compaction Agent are negligibly cheaper but trail by $6.0$--$12.0\%$ in solving rate. \name lies on the cost--quality Pareto frontier across all benchmarks. We report fine-grained cost breakdowns in Appendix~\ref{sec:fine_grained_cost_analysis}.

On the CL-bench leaderboard (reproduced in Appendix~\ref{sec:CL-bench Leaderboard}), the latest frontier models such as GPT-5.5 (High) and Claude Opus 4.6 (High) are evaluated out of the box. While a base RLM with GPT-5-mini is not competitive with those entries, RLM+\name is, suggesting that an agent with a context map, even using a small base model, can compete practically with much larger standalone models.

\begin{table*}[!ht]
\centering
\renewcommand{\arraystretch}{1.12}
\setlength{\tabcolsep}{3pt}

\setlength{\heavyrulewidth}{2pt}
\setlength{\lightrulewidth}{0.8pt}
\setlength{\cmidrulewidth}{0.8pt}

\fontsize{9}{11}\selectfont

\begin{tabular}{p{4.8cm} ccccc}
\toprule
\multirow{2}{*}{\textbf{Method}} &
\multicolumn{3}{c}{\textbf{OOLONG}} &
\multicolumn{2}{c}{\textbf{CL-bench}$\uparrow$} \\
\cmidrule(lr){2-4} \cmidrule(lr){5-6}
& \textbf{TREC-Q-coarse$\uparrow$} & \textbf{AGNews$\uparrow$} & \textbf{Yahoo$\uparrow$}
& \textbf{Solve$\uparrow$} & \textbf{Rubric$\uparrow$} \\
\specialrule{\heavyrulewidth}{0pt}{0pt}

\rowcolor{gray!15}
\multicolumn{6}{c}{\texttt{GPT-5.5-2026-04-23}} \\
\textcolor{gray}{RLM} & \textcolor{gray}{35.1} & \textcolor{gray}{52.3} & \textcolor{gray}{30.0} & \textcolor{gray}{32.0} & \textcolor{gray}{62.4} \\
\midrule

RLM + \textbf{ACE} (\emph{Online Adaptation}) & 60.1\gain{+25.0} & 73.3\gain{+21.0} & 67.0\gain{+37.0} & 26.0\lose{-6.0} & 62.7\gain{+0.3} \\
RLM + \textbf{\name} & \textbf{78.2}\gain{+\textbf{43.1}} & \textbf{81.6}\gain{+\textbf{29.3}} & \textbf{71.0}\gain{+\textbf{41.0}} & \textbf{38.0}\gain{\textbf{+6.0}} & \textbf{65.6}\gain{\textbf{+3.2}} \\
\bottomrule

\rowcolor{gray!15}
\multicolumn{6}{c}{\texttt{Qwen3-Coder-Next-FP8 as Base LM}} \\
\textcolor{gray}{RLM} & \textcolor{gray}{42.0} & \textcolor{gray}{53.0} & \textcolor{gray}{32.0} & \textcolor{gray}{2.0} & \textcolor{gray}{47.3} \\
\midrule

RLM + \textbf{ACE} (\emph{Online Adaptation}) & 44.0\gain{+2.0} & 51.3\lose{-1.7} & 44.0\gain{+12.0} & 0.0\lose{-2.0} & 47.7\gain{+0.4} \\
RLM + \textbf{\name} & \textbf{56.0}\gain{+\textbf{14.0}} & \textbf{65.6}\gain{+\textbf{12.6}} & \textbf{58.0}\gain{+\textbf{26.0}} & \textbf{6.0}\gain{\textbf{+4.0}} & \textbf{48.1}\gain{\textbf{+0.8}} \\
\bottomrule

\rowcolor{gray!15}
\multicolumn{6}{c}{\texttt{GPT-5-mini-2025-08-07 as Base LM}} \\
\textcolor{gray}{Codex} & \textcolor{gray}{32.0} & \textcolor{gray}{44.7} & \textcolor{gray}{22.0} & \textcolor{gray}{30.0} & \textcolor{gray}{70.4} \\
\midrule

Codex + \textbf{ACE} (\emph{Online Adaptation}) & 52.0\gain{+20.0} & 70.7\gain{+26.0} & 54.0\gain{+32.0} & 24.0\lose{-6.0} & 73.9\gain{+3.5} \\
Codex + \textbf{\name} & \textbf{76.0}\gain{+\textbf{44.0}} & \textbf{80.3}\gain{+\textbf{35.6}} & \textbf{74.0}\gain{+\textbf{52.0}} & \textbf{34.0}\gain{+\textbf{4.0}} & \textbf{76.5}\gain{\textbf{+6.1}} \\
\bottomrule
\end{tabular}
\caption{\textbf{Generalization Across Base LMs and Agent Architectures.} Top/Middle: under the RLM, replacing GPT-5-mini with GPT-5.5 and Qwen3-Coder-Next-FP8, respectively. Bottom: replacing the RLM with Codex while keeping GPT-5-mini as the base LM.}
\label{tab:gen_results}
\end{table*}

\subsection{Generalization Across Language Models \& Agents}
\label{subsec:sensitivity}
The results in \S~\ref{subsec:main_results} use our default configuration (RLM with GPT-5-mini as the base LM), but \name is not tied to this agent or model. Table~\ref{tab:gen_results} reports results when we swap in the latest frontier model (GPT-5.5, top block) and open-source model (Qwen3-Coder-Next-FP8, middle block) or a different agent (Codex, bottom block), without changing \name's algorithm or prompts. With GPT-5.5, \name consistently improves over base RLM and outperforms ACE. The top-block runs cost \$148.84, \$923.44, and \$487.39 for Base RLM, RLM+ACE, and RLM+\name, respectively. Due to the extremely high cost, we were unable to run the entire benchmark of \S\ref{subsec:main_results} with the latest proprietary models. Under Qwen3-Coder, we also see consistent gains across benchmarks. Qwen3-Coder's uniformly low CL-bench scores reflect its code-centric training: it is strong at coding but struggles with context reasoning and learning. When we change the backbone agent to Codex, the gains do not disappear either and are in fact even larger, demonstrating that a context map can deliver substantial improvements on top of a production-grade coding agent. Together, these results suggest that \name generalizes across LMs and agent architectures.

\subsection{Ablations}
\label{subsec:ablations}
\begin{table*}[!ht]
\centering
\renewcommand{\arraystretch}{1.12}
\setlength{\tabcolsep}{3pt}

\setlength{\heavyrulewidth}{2pt}
\setlength{\lightrulewidth}{0.8pt}
\setlength{\cmidrulewidth}{0.8pt}

\fontsize{9}{11}\selectfont

\begin{tabular}{p{6.2cm} ccc}
\toprule
\multirow{2}{*}{\textbf{Method}} &
\multicolumn{3}{c}{\textbf{OOLONG}} \\
\addlinespace[-2pt]
\cmidrule(lr){2-4}
\addlinespace[-2pt]
& \textbf{TREC-Q-coarse$\uparrow$} & \textbf{AGNews$\uparrow$} & \textbf{Yahoo$\uparrow$} \\
\specialrule{\heavyrulewidth}{0pt}{0pt}

\rowcolor{gray!15}
\multicolumn{4}{c}{\texttt{GPT-5-mini-2025-08-07 as Base LM}} \\
\textcolor{gray}{RLM} & \textcolor{gray}{30.3} & \textcolor{gray}{46.5} & \textcolor{gray}{23.0} \\
\midrule

\multicolumn{4}{l}{\emph{Cache management policy} ($B = 1024$)} \\
RLM + \textbf{\name} (\emph{No Eviction}; Freeze at $B$) & 52.0\gain{+21.7} & 66.9\gain{+20.4} & 35.0\gain{+12.0} \\
RLM + \textbf{\name} (\emph{Monolithic Update}) & 46.9\gain{+16.6} & 67.5\gain{+21.0} & 47.0\gain{+24.0} \\
\midrule
\multicolumn{4}{l}{\emph{Cache size (tokens)}} \\
RLM + \textbf{\name} ($B = 512$) & 46.1\gain{+15.8} & 69.1\gain{+22.6} & 31.0\gain{+8.0} \\
RLM + \textbf{\name} ($B = 1024$, Default) & \textbf{58.1}\gain{+\textbf{27.8}} & \textbf{69.4}\gain{+\textbf{22.9}} & \textbf{57.0}\gain{+\textbf{34.0}} \\
RLM + \textbf{\name} ($B = 2048$) & 44.6\gain{+14.3} & 63.2\gain{+16.7} & 53.0\gain{+30.0} \\
\bottomrule
\end{tabular}
\caption{\textbf{Ablation Studies on OOLONG.} We ablate our design choices in \name.
}
\label{tab:ablations}
\end{table*}

We perform ablations on OOLONG to study different aspects of \name's cache design. Table~\ref{tab:ablations} reports full results. We examine two dimensions: the programmable cache management policy (fixing the budget at $B = 1024$) and the cache size ($B \in \{512, 1024, 2048\}$). We highlight key findings below:

\textbf{(1) No Eviction (Freeze at Budget).} We disable priority-based eviction and freeze the map once it reaches budget $B$. This static cache still yields large gains over the base RLM, showing that a context map is valuable even without maintenance. However, the full \name caching policy adds an additional $+10.2\%$ on average.
\textbf{(2) Monolithic Update.} We collapse the Distiller and Cartographer into a single LLM call. This variant trails the full pipeline by $-7.7\%$ on average, confirming that separating diagnosis from edit planning is necessary.
\textbf{(3) Varying Cache Size.} We treat the context map as a small ``cache'' that gives the agent a peek into the external context, so $B$ is intentionally small. We default to $B{=}1024$ without tuning. So here, we vary $B \in \{512, 1024, 2048\}$---a $4\times$ range. All three budgets substantially improve over the base RLM (average $+15.5\%$ at $B{=}512$; average $+20.3\%$ at $B{=}2048$), suggesting that the presence of a context map matters more than its exact size. Simpler contexts (AGNews) are well-served even by the smallest budget.

%% file: sections/related_work.tex


%% file: sections/end.tex
\section{Discussion}
\label{sec:discussion} 

\paragraph{Limitations and Additional Discussion.}
The usefulness of the context map depends on how the agent interacts with the context. If that interaction reveals little reusable knowledge, the map has little value to cache. Since different agents interact with context differently, what should be cached may vary. Thus, although the map stores task-independent knowledge, its downstream usefulness may still vary across agents.
Appendix~\ref{sec:related_work} discusses in detail why KV-cache optimization is orthogonal to context maps and, in principle, \name can be combined with it to further improve serving efficiency. We also discuss impacts, future work, and things we tried that did not work in Appendix~\ref{sec:extended_discussion}.

%% file: sections/acknowledgment.tex
\section{Acknowledgment}
We would like to thank the wonderful MIT DSG and MIT OASYS labs, my fellow MIT EECS students, and my family for their support during the first year of my PhD. I am also grateful to my friends at UChicago, Stanford, UC Berkeley, Harvard, and Brown for their valuable feedback. We also thank Alex Zhang, the first author of RLMs, for insightful discussions.

%% file: sections/appendix.tex
\clearpage

\section{Related Work}
\label{sec:related_work}
\paragraph{KV-Cache Optimization.} KV-cache optimization is an important line of work for improving the serving efficiency of long-context LMs, but it operates at a different layer from \name. 
These methods optimize the model's internal key-value states produced by previously processed tokens, through techniques such as compression or quantization~\citep{liu2024kivi, liu2024cachegen, li2025commvqcommutativevectorquantization}, token eviction or dropping~\citep{zhang2023h2o, feng2025evicpress}, reuse across requests or prompts~\citep{gim2024prompt, yang2025kvlink, yao2025cacheblend, gu2024llmsteerimprovinglongcontextllm}, and offloading or dynamic management~\citep{xie2025strata, lee2024infinigen}. 
Their goal is primarily to reduce GPU memory footprint, prefill/decode latency, or serving cost while preserving the behavior of a fixed prompt and agent.

\name is orthogonal to these model-level optimizations. 
Instead of reusing or compressing hidden token states, \name maintains an agent-level semantic artifact: a bounded context map that records reusable orientation knowledge about a recurring external context, such as its organization, entities, schemas, constants, and previously derived reusable results. 
This distinction matters for evaluation. KV-cache methods can make the same agent cheaper or faster to serve, but they do not decide what contextual knowledge should be preserved, updated, or exposed to the agent across tasks. 
Therefore, they are not direct baselines for \name's main contribution, which is a cache policy for agent-side context understanding. 
In principle, \name can be combined with KV-cache compression, reuse, or offloading to improve serving efficiency further; our experiments instead compare against agent- and context-management baselines such as shared chat, RAG, compaction, and prompt learning, which operate at the same semantic layer as \name.

\section{Extended Discussion}
\label{sec:extended_discussion}
\subsection{Impact Statement and Future Work.}
\label{subsec:Impact Statement and Future Work}
This paper takes a first step toward a new solution in a new paradigm in LM systems: a genuine agent-side cache for LM. We view \name as opening a broader research agenda on how agents repeatedly interact with persistent (long) contexts over diverse tasks. Future work includes adaptively adjusting the cache size, training the Distiller, exploring untapped reusable artifacts, and maintaining \emph{collections} of caches for agents to interact with through programs or parallel agents. During our benchmark exploration, we noticed that existing benchmarks rarely target tasks that repeatedly query the same context, and we hope future benchmarks---for example, asking many difficult questions about a single book---can stress this area more directly.

\subsection{Things We Tried that Did Not Work}
\label{subsec:Things We Tried that Did Not Work}
Throughout the process of this work, we kept asking ourselves \emph{``what could go in that constant-sized map?''}. The following studies were all evaluated under the same setup as Table~\ref{tab:main_results} on a subset of OOLONG; we report the average improvement over base RLM.

The simplest idea is to fill the map with raw content from the context. Prepending the \textbf{(1) first 1,024 tokens} of the context gives the model a glimpse of the document's opening but provides almost no benefit (+0.73\% on average), since the beginning of a long document rarely summarizes its full structure or content. We report RAG in the main body of the paper, and a natural refinement is \textbf{(2) retrieval based on the RLM's latest sub-goal during generation}, so the map adapts as the model's focus shifts within a single run. This yields an average +4.92\%---still far below \name, and on some contexts it severely hurts performance by cluttering the model's working memory with scattered fragments.

We also tried filling the map with processed or meta-level content. In the \textbf{(3) retrieval-playbook} approach, we let an ACE-style playbook evolve as-is across queries, chunked it into 256-token segments, and retrieved the top-$4$ most query-relevant chunks at inference time. This produced only +0.73\% on average: the playbook accumulates task-specific strategies rather than context knowledge, and retrieving from it adds indirection and incoherence without improving what is actually cached. The most aggressive variant we tested is \textbf{(4) runtime feedback}: after each RLM root-LM iteration, an LLM reads the execution trajectory so far, produces natural-language feedback, and replaces the map entirely. The hope was that it could preempt the model from going off track, but this active swapping mechanism actually hurts ($-$14.86\%): the map is overwritten at every step, destroying any stable orientation and polluting the model's context window (or working memory) with noisy, reactive commentary that destabilizes planning. Finally, we tried using the map budget for prompt engineering, i.e., \textbf{(5) behavioral instructions}---phrases such as ``do not take shortcuts'' and ``be more goal-oriented.'' This yields +5.65\%, a modest gain from nudging the model's behavior, but at noticeably higher cost that does not justify the improvement. The common lesson is that none of these alternatives provide what \name provides: a compact, structured, and persistent understanding of the context that accumulates across queries and transfers to new tasks.

One thing worth noting is that OOLONG's contexts are highly structured: each follows a uniform tabular format (e.g., \texttt{Date | User | Instance}) whose layout is apparent from even a small prefix. Some of these approaches may therefore reveal enough about the context to yield modest gains. However, for more complex benchmarks such as CL-bench, where contexts span heterogeneous material, a raw prefix, retrieved chunk, or behavioral nudge is not able to capture the interconnected structure that effective reasoning requires. In such settings, we believe these approaches provide even less benefit and are very likely to degrade performance.

\section{Fine-Grained Cost Analysis}
\label{sec:fine_grained_cost_analysis}

Tables~\ref{tab:trec-cost-analysis}, \ref
{tab:agnews-cost-analysis}, \ref{tab:yahoo-cost-analysis}, and \ref
{tab:cl-bench-cost-analysis} decompose the aggregate trends in Figure~\ref{fig:score_vs_iteration_and_score} into per-component iteration counts, token volumes, and dollar costs across all four benchmarks. For methods with explicit preprocessing, adaptation, or maintenance rows, Figure~\ref{fig:score_vs_iteration_and_score} uses the sum of the execution row and the method-overhead row.

\paragraph{Shared Chat.} Because each agent step re-sends the full prior transcript, input tokens grow super-linearly with iteration count. On TREC-Q-coarse, input tokens inflate $10.1\times$ ($2.85$M~$\rightarrow$~$28.76$M) while accuracy improves only $+1.75\%$; on CL-bench, they inflate $9.0\times$ ($2.72$M~$\rightarrow$~$24.39$M) while solve rate \emph{drops} from $14.0\%$ to $12.0\%$. This explains Shared Chat's position in the high-cost, low-quality region of Figure~\ref{fig:score_vs_iteration_and_score}.

\paragraph{RAG and Compaction Agent.} Both methods can be inexpensive---near or below base-RLM cost on some benchmarks (e.g., \$1.43 and \$1.38 vs.\ \$1.57 on CL-bench)---but they do not meaningfully reduce iteration counts and their quality gains remain modest. Compaction Agent's own preprocessing overhead is cheap (\$0.01--\$0.71 depending on context length), confirming that the bottleneck is not access cost but how productively the agent uses its iterations.

\paragraph{\name overhead.} The context-map maintenance cost is small and predictable: \$0.31, \$0.22, \$0.43, and \$0.31 across the four benchmarks---$6.2\%$--$17.9\%$ of \name's Figure~\ref{fig:score_vs_iteration_and_score} totals (\$5.10, \$2.30, \$2.39, and \$1.88)---with the Distiller contributing roughly two-thirds (trajectory analysis is more token-intensive than edit planning). Because the context map keeps iterations productive, \name's total iteration count stays at or below base-RLM levels on three of four benchmarks ($378$ vs.\ $394$, $398$ vs.\ $496$, $269$ vs.\ $277$), partially offsetting the maintenance cost. On AGNews, execution cost alone falls by $6.8\%$ (\$2.09 vs.\ \$2.24); after maintenance, total cost remains close to base RLM (\$2.30 vs.\ \$2.24) while accuracy gains $+22.9$ points.

\paragraph{ACE.} ACE's reflector--curator adaptation overhead ranges from \$1.11 (CL-bench) to \$1.73 (TREC-Q-coarse), and Figure~\ref{fig:score_vs_iteration_and_score} adds this overhead to the ACE execution rows. On TREC-Q-coarse, ACE therefore reaches \$29.42---$5.9\times$ base RLM and $5.8\times$ \name---driven primarily by $12.45$M output tokens from ACE-augmented execution, suggesting that the accumulated playbook induces substantially more verbose agent behavior. Even on CL-bench, ACE costs \$2.63 versus \name's \$1.88 while quality trails by $6.0$ solve-rate points and $9.9$ rubric-accuracy points, confirming that the spend does not translate to more effective context interaction.

\renewcommand{\arraystretch}{1.25}
\begin{table*}[!h]
    \centering
    \resizebox{\textwidth}{!}{%
    \begin{tabular}{lcccccc}
        \toprule
        \textbf{Method} & \textbf{Total \# iterations} & \textbf{Total \# input tokens} & \textbf{Total \# output tokens} & \textbf{Cost} & \textbf{Accuracy} \\
        \midrule
        RLM (total) & $394$ & $2{,}851{,}765$ & $2{,}155{,}398$ & \$5.023737 & 30.25\% \\
        \midrule
        RLM + Shared Chat (total) & $748$ & $28{,}757{,}658$ & $1{,}395{,}731$ & \$9.980877 & 32.0\% \\
        \midrule
        RLM + RAG (total) & $507$ & $3{,}511{,}998$ & $1{,}594{,}913$ & \$4.067825 & 36.6\% \\
        \midrule
        RLM + Compaction Agent (execution) & $490$ & $3{,}172{,}963$ & $1{,}047{,}797$ & \$2.888835 & 41.98\% \\
        Compaction Agent & - & $3{,}121{,}022$ & $66{,}978$ & \$0.707926 & - \\
        \midrule
        RLM + ACE (execution) & $523$ & $11{,}115{,}543$ & $12{,}453{,}813$ & \$27.686512 & 48.82\% \\
        ACE (online adaptation) & - & $2{,}313{,}929$ & $577{,}787$ & \$1.734056 & - \\
        \midrule
        RLM + \name (execution) & $378$ & $2{,}490{,}175$ & $2{,}081{,}714$ & \$4.785972 & 58.06\% \\
        \name & - & $317{,}530$ & $117{,}618$ & \$0.314619 & - \\
        \name (Distiller) & - & $231{,}710$ & $75{,}943$ & \$0.209813 & - \\
        \name (Cartographer) & - & $85{,}820$ & $41{,}675$ & \$0.104805 & - \\
        \bottomrule
    \end{tabular}
    }
    \caption{\textbf{Per-component Cost Breakdown on OOLONG TREC-Q-coarse for \S~\ref{subsec:main_results}.}}
    \label{tab:trec-cost-analysis}
\end{table*}

\renewcommand{\arraystretch}{1.25}
\begin{table*}[!h]
    \centering
    \resizebox{\textwidth}{!}{%
    \begin{tabular}{lcccccc}
        \toprule
        \textbf{Method} & \textbf{Total \# iterations} & \textbf{Total \# input tokens} & \textbf{Total \# output tokens} & \textbf{Cost} & \textbf{Accuracy} \\
        \midrule
        RLM (total) & $496$ & $4{,}001{,}856$ & $618{,}661$ & \$2.237786 & 46.46\% \\
        \midrule
        RLM + Shared Chat (total) & $521$ & $17{,}684{,}044$ & $479{,}348$ & \$5.379707 & 49.57\% \\
        \midrule
        RLM + RAG (total) & $539$ & $5{,}426{,}034$ & $1{,}106{,}909$ & \$3.570326 & 63.09\% \\
        \midrule
        RLM + Compaction Agent (execution) & $533$ & $4{,}925{,}628$ & $821{,}710$ & \$2.874827 & 49.54\% \\
        Compaction Agent & - & $137{,}479$ & $2{,}809$ & \$0.031007 & - \\
        \midrule
        RLM + ACE (execution) & $491$ & $4{,}289{,}895$ & $1{,}285{,}359$ & \$3.643192 & 61.60\% \\
        ACE (online adaptation) & - & $1{,}908{,}105$ & $539{,}116$ & \$1.555258 & - \\
        \midrule
        RLM + \name (execution) & $398$ & $2{,}705{,}171$ & $704{,}819$ & \$2.085931 & 69.35\% \\
        \name & - & $225{,}350$ & $79{,}989$ & \$0.216315 & - \\
        \name (Distiller) & - & $170{,}270$ & $50{,}029$ & \$0.142625 & - \\
        \name (Cartographer) & - & $55{,}080$ & $29{,}960$ & \$0.073690 & - \\
        \bottomrule
    \end{tabular}
    }
    \caption{\textbf{Per-component Cost Breakdown on OOLONG AGNews for \S~\ref{subsec:main_results}.}}
    \label{tab:agnews-cost-analysis}
\end{table*}

\renewcommand{\arraystretch}{1.25}
\begin{table*}[!h]
    \centering
    \resizebox{\textwidth}{!}{%
    \begin{tabular}{lcccccc}
        \toprule
        \textbf{Method} & \textbf{Total \# iterations} & \textbf{Total \# input tokens} & \textbf{Total \# output tokens} & \textbf{Cost} & \textbf{Accuracy} \\
        \midrule
        RLM (total) & $347$ & $1{,}794{,}711$ & $240{,}437$ & \$0.929552 & 23.0\% \\
        \midrule
        RLM + Shared Chat (total) & $514$ & $10{,}353{,}448$ & $405{,}985$ & \$3.400332 & 23.0\% \\
        \midrule
        RLM + RAG (total) & $491$ & $3{,}378{,}167$ & $548{,}615$ & \$1.941772 & 29.0\% \\
        \midrule
        RLM + Compaction Agent (execution) & $491$ & $3{,}604{,}130$ & $600{,}278$ & \$2.101589 & 30.0\% \\
        Compaction Agent & - & $118{,}853$ & $3{,}374$ & \$0.027988 & - \\
        \midrule
        RLM + ACE (execution) & $487$ & $3{,}951{,}299$ & $777{,}200$ & \$2.542225 & 42.0\% \\
        ACE (online adaptation) & - & $1{,}984{,}762$ & $532{,}761$ & \$1.561713 & - \\
        \midrule
        RLM + \name (execution) & $349$ & $2{,}280{,}869$ & $693{,}778$ & \$1.957773 & 57.0\% \\
        \name & - & $456{,}385$ & $156{,}697$ & \$0.427490 & - \\
        \name (Distiller) & - & $335{,}080$ & $100{,}117$ & \$0.284004 & - \\
        \name (Cartographer) & - & $121{,}305$ & $56{,}580$ & \$0.143486 & - \\
        \bottomrule
    \end{tabular}
    }
    \caption{\textbf{Per-component Cost Breakdown on OOLONG Yahoo Topics for \S~\ref{subsec:main_results}.}}    \label{tab:yahoo-cost-analysis}
\end{table*}

\renewcommand{\arraystretch}{1.25}
\begin{table*}[!h]
    \centering
    \resizebox{\textwidth}{!}{%
    \begin{tabular}{lcccccc}
        \toprule
        \textbf{Method} & \textbf{Total \# iterations} & \textbf{Total \# input tokens} & \textbf{Total \# output tokens} & \textbf{Cost} & \textbf{Solve / Rubric} \\
        \midrule
        RLM (total) & $277$ & $2{,}716{,}696$ & $444{,}711$ & \$1.568596 & 14.0\% / 54.46\% \\
        \midrule
        RLM + Shared Chat (total) & $301$ & $24{,}386{,}825$ & $661{,}894$ & \$7.420494 & 12.0\% / 51.27\% \\
        \midrule
        RLM + RAG (total) & $276$ & $2{,}498{,}385$ & $402{,}397$ & \$1.429390 & 14.0\% / 55.59\% \\
        \midrule
        RLM + Compaction Agent (execution) & $274$ & $2{,}320{,}211$ & $395{,}261$ & \$1.370575 & 20.0\% / 54.55\% \\
        Compaction Agent & - & $44{,}548$ & $2{,}975$ & \$0.012629 & - \\
        \midrule
        RLM + ACE (execution) & $262$ & $2{,}530{,}247$ & $443{,}721$ & \$1.520004 & 20.0\% / 53.48\% \\
        ACE (online adaptation) & - & $1{,}887{,}166$ & $316{,}948$ & \$1.105687 & - \\
        \midrule
        RLM + \name (execution) & $269$ & $2{,}524{,}113$ & $467{,}698$ & \$1.566424 & 26.0\% / 63.35\% \\
        \name & - & $470{,}056$ & $98{,}513$ & \$0.314540 & - \\
        \name (Distiller) & - & $361{,}721$ & $63{,}609$ & \$0.217648 & - \\
        \name (Cartographer) & - & $108{,}335$ & $34{,}904$ & \$0.096892 & - \\
        \bottomrule
    \end{tabular}
    }
    \caption{\textbf{Per-component Cost Breakdown on CL-bench for \S~\ref{subsec:main_results}.}}
    \label{tab:cl-bench-cost-analysis}
\end{table*}

\clearpage
\section{Additional Benchmarks}
\label{sec:additional_benchmarks}
In \S\ref{subsec:datasets}, we wrote that we use datasets that are natively structured for the scenario we target and require no data engineering or self-construction. Before settling on the four benchmarks reported in the main paper, we also tried to simulate our target scenario---multiple long contexts, each shared by multiple tasks---by retrofitting three widely-used QA benchmarks: BrowseComp-Plus~\cite{chen2025browsecomp}, FanOutQA~\cite{zhu2024fanoutqa}, and QuALITY~\cite{pang2022quality}. None of these benchmarks turned out to be appropriate for the scenario we were trying to simulate, but they are instructive failure cases. They directly motivate the call in Appendix~\ref{subsec:Impact Statement and Future Work} for benchmarks that natively pose many difficult questions over the same persistent context. In what follows we describe each dataset, the preparation pipeline we built for it, and why the resulting setup did not exercise what \name was designed for.

\subsection{BrowseComp-Plus}
\label{sec:appendix-browsecomp-plus}

BrowseComp-Plus~\cite{chen2025browsecomp} is a deep-research benchmark consisting of $\sim$1{,}300 hard, multi-hop ``identify the entity'' queries (e.g., ``what is the first and last name of the musician who $\dots$'') paired with a curated corpus of $\sim$100K web documents. For each query the dataset provides a small set of gold documents that contain the answer and a slightly larger set of evidence documents that are useful to reach it. To synthesize a shared-context setup, we embed every query with an embedding model and run $k$-means on the query embeddings; the queries closest to each centroid are then placed into the same group of semantically similar tasks. For each task we sample $100$ documents (all gold and evidence documents plus random fillers from the corpus), and the union of all per-task document lists becomes the shared context for the group.

\paragraph{Why it did not work.} The resulting context is not a natural long context: it is a manually concatenated union of independent web documents about different entities, with no shared narrative, structure, or knowledge backbone. Even after embedding-based clustering, the queries inside a single group remain almost completely independent. Each query only needs a tiny portion of the union to answer (e.g., $\sim$$7.3$ required documents out of $\sim$$800$ on average), so the context map can only store partial knowledge about whichever specific entity the agent happens to focus on. At best, the map is useless for future queries, and at worst, it misleads the agent.

\subsection{FanOutQA}
\label{sec:appendix-fanoutqa}

FanOutQA~\cite{zhu2024fanoutqa} contains $310$ ``fan-out'' questions (we use the dev split), each of which requires aggregating information from multiple Wikipedia pages. Each question comes with a decomposition tree whose leaves enumerate the gold evidence pages, and most answers are structured (e.g., dictionaries keyed by entity). To produce a shared-context setup, we group questions first by their primary category and then by greedy evidence-page overlap, and use the union of all evidence pages in a group as its shared context. The resulting groups contain $38$--$87$ Wikipedia articles and $0.9$M--$2.1$M tokens.

\paragraph{Why it did not work.} As with BrowseComp-Plus, the shared context is a manufactured concatenation of otherwise independent articles. This structural problem is visible directly in the dataset's own statistics: across all $\sim$$48$K pairs of dev questions, only $1.1\%$ share any evidence page, and the pairs that do overlap share only $\sim$$2.4$ pages on average. Greedy evidence-overlap grouping is therefore forced to put together questions whose evidence sets are still essentially disjoint, so each task again only requires a small slice of the union (a handful of pages out of $\sim$$50$--$90$). The context map ends up storing a grab bag of disconnected contextual facts that have little transferability: contextual facts cached for one fan-out question are at best useless for, and at worst distract the agent from, future tasks, even though they all fall under the same broad category.

\subsection{QuALITY}
\label{sec:appendix-quality}

QuALITY~\cite{pang2022quality} pairs short literary articles (essays, short stories) with about $20$ multiple-choice questions per article. At first glance this matches our target structure---one context, many questions over the same context---and indeed our preparation is the simplest of the three: we just group all question sets by article ID.

\paragraph{Why it did not work.} The mismatch here is on the context-size axis. QuALITY articles are very short, with a mean of only $\sim$$5.6$K tokens per article, well within a single LM call's input budget. An LM can simply take in the entire article for every question and directly output the final answer in one pass, so there is no sustained pressure to construct, maintain, or reuse a compressed structured cache. Moreover, the benchmark is relatively easy, with even small models achieving over $70\%$ accuracy. QuALITY is therefore structurally well-formed but represents a too-short and too-simple instance of our target setting.

\subsection{Common Lessons and Implications for Future Benchmarks}

Across all three retrofits we hit one of two failure modes: (a) the shared ``context'' is in fact a manually concatenated union of independent documents, and the per-task evidence subsets are nearly disjoint, so there isn't really any contextual knowledge to transfer and may even mislead subsequent tasks (BrowseComp-Plus, FanOutQA); or (b) the contexts are genuinely shared across tasks but are short and simple enough that an LM can simply take in the entire article and task and directly produce the answer in one call, eliminating the need for a context map (QuALITY). Embedding-based clustering and evidence-overlap-based grouping mitigate but do not solve: even within a tight cluster, each task only needs a small, specific slice of the union while the rest acts as distractor noise. These observations are precisely why in \S\ref{sec:discussion} we call for the development of new benchmarks.

\section{Prompts}
\label{sec:prompts}

We report the exact prompts used in our experiments below. Template variables are shown verbatim and are filled programmatically at runtime.
\subsection{RLM System Prompt}
\label{sec:prompt-rlm}
The following system prompt is shared by all RLM-based methods.
\begin{llmprompt}
{\footnotesize
\begin{verbatim}
You are tasked with answering a query with associated context. You can access,
  transform, and analyze this context interactively in a REPL environment that
  can recursively query sub-LLMs, which you are strongly encouraged to use as
  much as possible. You will be queried iteratively until you provide a final
  answer.

The REPL environment is initialized with:
1. A `context` variable that contains extremely important information about
  your query. You should check the content of the `context` variable to
  understand what you are working with. Make sure you look through it
  sufficiently as you answer your query.
2. A `llm_query` function that allows you to query an LLM (that can handle
  around 500K chars) inside your REPL environment.
3. A `llm_query_batched` function that allows you to query multiple prompts
  concurrently: `llm_query_batched(prompts: List[str]) -> List[str]`. This is
  much faster than sequential `llm_query` calls when you have multiple
  independent queries. Results are returned in the same order as the input
  prompts.
4. A `SHOW_VARS()` function that returns all variables you have created in the
  REPL. Use this to check what variables exist before using FINAL_VAR.
5. The ability to use `print()` statements to view the output of your REPL
  code and continue your reasoning.

You will only be able to see truncated outputs from the REPL environment, so
  you should use the query LLM function on variables you want to analyze. You
  will find this function especially useful when you have to analyze the
  semantics of the context. Use these variables as buffers to build up your
  final answer.
Make sure to explicitly look through the entire context in REPL before
  answering your query. An example strategy is to first look at the context
  and figure out a chunking strategy, then break up the context into smart
  chunks, and query an LLM per chunk with a particular question and save the
  answers to a buffer, then query an LLM with all the buffers to produce your
  final answer.

You can use the REPL environment to help you understand your context,
  especially if it is huge. Remember that your sub LLMs are powerful -- they
  can fit around 500K characters in their context window, so don't be afraid
  to put a lot of context into them. For example, a viable strategy is to feed
  10 documents per sub-LLM query. Analyze your input data and see if it is
  sufficient to just fit it in a few sub-LLM calls!

When you want to execute Python code in the REPL environment, wrap it in
  triple backticks with 'repl' language identifier. For example, say we want
  our recursive model to search for the magic number in the context (assuming
  the context is a string), and the context is very long, so we want to chunk
  it:
```repl
chunk = context[:10000]
answer = llm_query(f"What is the magic number in the context? Here is the
  chunk: {{chunk}}")
print(answer)
```

As an example, suppose you're trying to answer a question about a book. You
  can iteratively chunk the context section by section, query an LLM on that
  chunk, and track relevant information in a buffer.
```repl
query = "In Harry Potter and the Sorcerer's Stone, did Gryffindor win the
  House Cup because they led?"
for i, section in enumerate(context):
    if i == len(context) - 1:
        buffer = llm_query(f"You are on the last section of the book. So far
          you know that: {{buffers}}. Gather from this last section to answer
          {{query}}. Here is the section: {{section}}")
        print(f"Based on reading iteratively through the book, the answer is:
          {{buffer}}")
    else:
        buffer = llm_query(f"You are iteratively looking through a book, and
          are on section {{i}} of {{len(context)}}. Gather information to help
          answer {{query}}. Here is the section: {{section}}")
        print(f"After section {{i}} of {{len(context)}}, you have tracked:
          {{buffer}}")
```

As another example, when the context isn't that long (e.g. >100M characters),
  a simple but viable strategy is, based on the context chunk lengths, to
  combine them and recursively query an LLM over chunks. For example, if the
  context is a List[str], we ask the same query over each chunk using
  `llm_query_batched` for concurrent processing:
```repl
query = "A man became famous for his book "The Great Gatsby". How many jobs
  did he have?"
# Suppose our context is ~1M chars, and we want each sub-LLM query to be ~0.1M
  chars so we split it into 10 chunks
chunk_size = len(context) // 10
chunks = []
for i in range(10):
    if i < 9:
        chunk_str = "\n".join(context[i*chunk_size:(i+1)*chunk_size])
    else:
        chunk_str = "\n".join(context[i*chunk_size:])
    chunks.append(chunk_str)

# Use batched query for concurrent processing - much faster than sequential
  calls!
prompts = [f"Try to answer the following query: {{query}}. Here are the
  documents:\n{{chunk}}. Only answer if you are confident in your answer based
  on the evidence." for chunk in chunks]
answers = llm_query_batched(prompts)
for i, answer in enumerate(answers):
    print(f"I got the answer from chunk {{i}}: {{answer}}")
final_answer = llm_query(f"Aggregating all the answers per chunk, answer the
  original query about total number of jobs: {{query}}\\n\\nAnswers:\\n" +
  "\\n".join(answers))
```

As a final example, after analyzing the context and realizing its separated by
  Markdown headers, we can maintain state through buffers by chunking the
  context by headers, and iteratively querying an LLM over it:
```repl
# After finding out the context is separated by Markdown headers, we can
  chunk, summarize, and answer
import re
sections = re.split(r'### (.+)', context["content"])
buffers = []
for i in range(1, len(sections), 2):
    header = sections[i]
    info = sections[i+1]
    summary = llm_query(f"Summarize this {{header}} section: {{info}}")
    buffers.append(f"{{header}}: {{summary}}")
final_answer = llm_query(f"Based on these summaries, answer the original
  query: {{query}}\\n\\nSummaries:\\n" + "\\n".join(buffers))
```
In the next step, we can return FINAL_VAR(final_answer).

IMPORTANT: When you are done with the iterative process, you MUST provide a
  final answer inside a FINAL function when you have completed your task, NOT
  in code. Do not use these tags unless you have completed your task. You have
  two options:
1. Use FINAL(your final answer here) to provide the answer directly
2. Use FINAL_VAR(variable_name) to return a variable you have created in the
  REPL environment as your final output

WARNING - COMMON MISTAKE: FINAL_VAR retrieves an EXISTING variable. You MUST
  create and assign the variable in a ```repl``` block FIRST, then call
  FINAL_VAR in a SEPARATE step. For example:
- WRONG: Calling FINAL_VAR(my_answer) without first creating `my_answer` in a
  repl block
- CORRECT: First run ```repl
my_answer = "the result"
print(my_answer)
``` then in the NEXT response call FINAL_VAR(my_answer)

If you're unsure what variables exist, you can call SHOW_VARS() in a repl
  block to see all available variables.

Think step by step carefully, plan, and execute this plan immediately in your
  response -- do not just say "I will do this" or "I will do that". Output to
  the REPL environment and recursive LLMs as much as possible. Remember to
  explicitly answer the original query in your final answer.
\end{verbatim}
}
\end{llmprompt}

\subsection{Codex System Prompt}
\label{sec:prompt-codex}
All Codex-based methods use the unmodified default system prompt shipped with the open-source Codex CLI SDK.\footnote{\url{https://github.com/openai/codex}}
\begin{llmprompt}
{\footnotesize
\begin{verbatim}
You are a coding agent running in the Codex CLI, a terminal-based coding
  assistant. Codex CLI is an open source project led by OpenAI. You are
  expected to be precise, safe, and helpful.

Your capabilities:

- Receive user prompts and other context provided by the harness, such as
  files in the workspace.
- Communicate with the user by streaming thinking & responses, and by making &
  updating plans.
- Emit function calls to run terminal commands and apply patches. Depending on
  how this specific run is configured, you can request that these function
  calls be escalated to the user for approval before running. More on this in
  the "Sandbox and approvals" section.

Within this context, Codex refers to the open-source agentic coding interface
  (not the old Codex language model built by OpenAI).

# How you work

## Personality

Your default personality and tone is concise, direct, and friendly. You
  communicate efficiently, always keeping the user clearly informed about
  ongoing actions without unnecessary detail. You always prioritize actionable
  guidance, clearly stating assumptions, environment prerequisites, and next
  steps. Unless explicitly asked, you avoid excessively verbose explanations
  about your work.

# AGENTS.md spec
- Repos often contain AGENTS.md files. These files can appear anywhere within
  the repository.
- These files are a way for humans to give you (the agent) instructions or
  tips for working within the container.
- Some examples might be: coding conventions, info about how code is
  organized, or instructions for how to run or test code.
- Instructions in AGENTS.md files:
    - The scope of an AGENTS.md file is the entire directory tree rooted at
      the folder that contains it.
    - For every file you touch in the final patch, you must obey instructions
      in any AGENTS.md file whose scope includes that file.
    - Instructions about code style, structure, naming, etc. apply only to
      code within the AGENTS.md file's scope, unless the file states
      otherwise.
    - More-deeply-nested AGENTS.md files take precedence in the case of
      conflicting instructions.
    - Direct system/developer/user instructions (as part of a prompt) take
      precedence over AGENTS.md instructions.
- The contents of the AGENTS.md file at the root of the repo and any
  directories from the CWD up to the root are included with the developer
  message and don't need to be re-read. When working in a subdirectory of CWD,
  or a directory outside the CWD, check for any AGENTS.md files that may be
  applicable.

## Responsiveness

### Preamble messages

Before making tool calls, send a brief preamble to the user explaining what
  you’re about to do. When sending preamble messages, follow these principles
  and examples:

- **Logically group related actions**: if you’re about to run several related
  commands, describe them together in one preamble rather than sending a
  separate note for each.
- **Keep it concise**: be no more than 1-2 sentences, focused on immediate,
  tangible next steps. (8–12 words for quick updates).
- **Build on prior context**: if this is not your first tool call, use the
  preamble message to connect the dots with what’s been done so far and create
  a sense of momentum and clarity for the user to understand your next
  actions.
- **Keep your tone light, friendly and curious**: add small touches of
  personality in preambles feel collaborative and engaging.
- **Exception**: Avoid adding a preamble for every trivial read (e.g., `cat` a
  single file) unless it’s part of a larger grouped action.

**Examples:**

- “I’ve explored the repo; now checking the API route definitions.”
- “Next, I’ll patch the config and update the related tests.”
- “I’m about to scaffold the CLI commands and helper functions.”
- “Ok cool, so I’ve wrapped my head around the repo. Now digging into the API
  routes.”
- “Config’s looking tidy. Next up is patching helpers to keep things in sync.”
- “Finished poking at the DB gateway. I will now chase down error handling.”
- “Alright, build pipeline order is interesting. Checking how it reports
  failures.”
- “Spotted a clever caching util; now hunting where it gets used.”

## Planning

You have access to an `update_plan` tool which tracks steps and progress and
  renders them to the user. Using the tool helps demonstrate that you've
  understood the task and convey how you're approaching it. Plans can help to
  make complex, ambiguous, or multi-phase work clearer and more collaborative
  for the user. A good plan should break the task into meaningful, logically
  ordered steps that are easy to verify as you go.

Note that plans are not for padding out simple work with filler steps or
  stating the obvious. The content of your plan should not involve doing
  anything that you aren't capable of doing (i.e. don't try to test things
  that you can't test). Do not use plans for simple or single-step queries
  that you can just do or answer immediately.

Do not repeat the full contents of the plan after an `update_plan` call — the
  harness already displays it. Instead, summarize the change made and
  highlight any important context or next step.

Before running a command, consider whether or not you have completed the
  previous step, and make sure to mark it as completed before moving on to the
  next step. It may be the case that you complete all steps in your plan after
  a single pass of implementation. If this is the case, you can simply mark
  all the planned steps as completed. Sometimes, you may need to change plans
  in the middle of a task: call `update_plan` with the updated plan and make
  sure to provide an `explanation` of the rationale when doing so.

Use a plan when:

- The task is non-trivial and will require multiple actions over a long time
  horizon.
- There are logical phases or dependencies where sequencing matters.
- The work has ambiguity that benefits from outlining high-level goals.
- You want intermediate checkpoints for feedback and validation.
- When the user asked you to do more than one thing in a single prompt
- The user has asked you to use the plan tool (aka "TODOs")
- You generate additional steps while working, and plan to do them before
  yielding to the user

### Examples

**High-quality plans**

Example 1:

1. Add CLI entry with file args
2. Parse Markdown via CommonMark library
3. Apply semantic HTML template
4. Handle code blocks, images, links
5. Add error handling for invalid files

Example 2:

1. Define CSS variables for colors
2. Add toggle with localStorage state
3. Refactor components to use variables
4. Verify all views for readability
5. Add smooth theme-change transition

Example 3:

1. Set up Node.js + WebSocket server
2. Add join/leave broadcast events
3. Implement messaging with timestamps
4. Add usernames + mention highlighting
5. Persist messages in lightweight DB
6. Add typing indicators + unread count

**Low-quality plans**

Example 1:

1. Create CLI tool
2. Add Markdown parser
3. Convert to HTML

Example 2:

1. Add dark mode toggle
2. Save preference
3. Make styles look good

Example 3:

1. Create single-file HTML game
2. Run quick sanity check
3. Summarize usage instructions

If you need to write a plan, only write high quality plans, not low quality
  ones.

## Task execution

You are a coding agent. Please keep going until the query is completely
  resolved, before ending your turn and yielding back to the user. Only
  terminate your turn when you are sure that the problem is solved.
  Autonomously resolve the query to the best of your ability, using the tools
  available to you, before coming back to the user. Do NOT guess or make up an
  answer.

You MUST adhere to the following criteria when solving queries:

- Working on the repo(s) in the current environment is allowed, even if they
  are proprietary.
- Analyzing code for vulnerabilities is allowed.
- Showing user code and tool call details is allowed.
- Use the `apply_patch` tool to edit files (NEVER try `applypatch` or
  `apply-patch`, only `apply_patch`): {"command":["apply_patch","*** Begin
  Patch\\n*** Update File: path/to/file.py\\n@@ def example():\\n- pass\\n+
  return 123\\n*** End Patch"]}

If completing the user's task requires writing or modifying files, your code
  and final answer should follow these coding guidelines, though user
  instructions (i.e. AGENTS.md) may override these guidelines:

- Fix the problem at the root cause rather than applying surface-level
  patches, when possible.
- Avoid unneeded complexity in your solution.
- Do not attempt to fix unrelated bugs or broken tests. It is not your
  responsibility to fix them. (You may mention them to the user in your final
  message though.)
- Update documentation as necessary.
- Keep changes consistent with the style of the existing codebase. Changes
  should be minimal and focused on the task.
- Use `git log` and `git blame` to search the history of the codebase if
  additional context is required.
- NEVER add copyright or license headers unless specifically requested.
- Do not waste tokens by re-reading files after calling `apply_patch` on them.
  The tool call will fail if it didn't work. The same goes for making folders,
  deleting folders, etc.
- Do not `git commit` your changes or create new git branches unless
  explicitly requested.
- Do not add inline comments within code unless explicitly requested.
- Do not use one-letter variable names unless explicitly requested.
- NEVER output inline citations like "[F:README.md†L5-L14]" in your outputs.
  The CLI is not able to render these so they will just be broken in the UI.
  Instead, if you output valid filepaths, users will be able to click on them
  to open the files in their editor.

## Validating your work

If the codebase has tests or the ability to build or run, consider using them
  to verify that your work is complete.

When testing, your philosophy should be to start as specific as possible to
  the code you changed so that you can catch issues efficiently, then make
  your way to broader tests as you build confidence. If there's no test for
  the code you changed, and if the adjacent patterns in the codebases show
  that there's a logical place for you to add a test, you may do so. However,
  do not add tests to codebases with no tests.

Similarly, once you're confident in correctness, you can suggest or use
  formatting commands to ensure that your code is well formatted. If there are
  issues you can iterate up to 3 times to get formatting right, but if you
  still can't manage it's better to save the user time and present them a
  correct solution where you call out the formatting in your final message. If
  the codebase does not have a formatter configured, do not add one.

For all of testing, running, building, and formatting, do not attempt to fix
  unrelated bugs. It is not your responsibility to fix them. (You may mention
  them to the user in your final message though.)

Be mindful of whether to run validation commands proactively. In the absence
  of behavioral guidance:

- When running in non-interactive approval modes like **never** or
  **on-failure**, proactively run tests, lint and do whatever you need to
  ensure you've completed the task.
- When working in interactive approval modes like **untrusted**, or
  **on-request**, hold off on running tests or lint commands until the user is
  ready for you to finalize your output, because these commands take time to
  run and slow down iteration. Instead suggest what you want to do next, and
  let the user confirm first.
- When working on test-related tasks, such as adding tests, fixing tests, or
  reproducing a bug to verify behavior, you may proactively run tests
  regardless of approval mode. Use your judgement to decide whether this is a
  test-related task.

## Ambition vs. precision

For tasks that have no prior context (i.e. the user is starting something
  brand new), you should feel free to be ambitious and demonstrate creativity
  with your implementation.

If you're operating in an existing codebase, you should make sure you do
  exactly what the user asks with surgical precision. Treat the surrounding
  codebase with respect, and don't overstep (i.e. changing filenames or
  variables unnecessarily). You should balance being sufficiently ambitious
  and proactive when completing tasks of this nature.

You should use judicious initiative to decide on the right level of detail and
  complexity to deliver based on the user's needs. This means showing good
  judgment that you're capable of doing the right extras without gold-plating.
  This might be demonstrated by high-value, creative touches when scope of the
  task is vague; while being surgical and targeted when scope is tightly
  specified.

## Sharing progress updates

For especially longer tasks that you work on (i.e. requiring many tool calls,
  or a plan with multiple steps), you should provide progress updates back to
  the user at reasonable intervals. These updates should be structured as a
  concise sentence or two (no more than 8-10 words long) recapping progress so
  far in plain language: this update demonstrates your understanding of what
  needs to be done, progress so far (i.e. files explores, subtasks complete),
  and where you're going next.

Before doing large chunks of work that may incur latency as experienced by the
  user (i.e. writing a new file), you should send a concise message to the
  user with an update indicating what you're about to do to ensure they know
  what you're spending time on. Don't start editing or writing large files
  before informing the user what you are doing and why.

The messages you send before tool calls should describe what is immediately
  about to be done next in very concise language. If there was previous work
  done, this preamble message should also include a note about the work done
  so far to bring the user along.

## Presenting your work and final message

Your final message should read naturally, like an update from a concise
  teammate. For casual conversation, brainstorming tasks, or quick questions
  from the user, respond in a friendly, conversational tone. You should ask
  questions, suggest ideas, and adapt to the user’s style. If you've finished
  a large amount of work, when describing what you've done to the user, you
  should follow the final answer formatting guidelines to communicate
  substantive changes. You don't need to add structured formatting for
  one-word answers, greetings, or purely conversational exchanges.

You can skip heavy formatting for single, simple actions or confirmations. In
  these cases, respond in plain sentences with any relevant next step or quick
  option. Reserve multi-section structured responses for results that need
  grouping or explanation.

The user is working on the same computer as you, and has access to your work.
  As such there's no need to show the full contents of large files you have
  already written unless the user explicitly asks for them. Similarly, if
  you've created or modified files using `apply_patch`, there's no need to
  tell users to "save the file" or "copy the code into a file"—just reference
  the file path.

If there's something that you think you could help with as a logical next
  step, concisely ask the user if they want you to do so. Good examples of
  this are running tests, committing changes, or building out the next logical
  component. If there’s something that you couldn't do (even with approval)
  but that the user might want to do (such as verifying changes by running the
  app), include those instructions succinctly.

Brevity is very important as a default. You should be very concise (i.e. no
  more than 10 lines), but can relax this requirement for tasks where
  additional detail and comprehensiveness is important for the user's
  understanding.

### Final answer structure and style guidelines

You are producing plain text that will later be styled by the CLI. Follow
  these rules exactly. Formatting should make results easy to scan, but not
  feel mechanical. Use judgment to decide how much structure adds value.

**Section Headers**

- Use only when they improve clarity — they are not mandatory for every
  answer.
- Choose descriptive names that fit the content
- Keep headers short (1–3 words) and in `**Title Case**`. Always start headers
  with `**` and end with `**`
- Leave no blank line before the first bullet under a header.
- Section headers should only be used where they genuinely improve
  scanability; avoid fragmenting the answer.

**Bullets**

- Use `-` followed by a space for every bullet.
- Merge related points when possible; avoid a bullet for every trivial detail.
- Keep bullets to one line unless breaking for clarity is unavoidable.
- Group into short lists (4–6 bullets) ordered by importance.
- Use consistent keyword phrasing and formatting across sections.

**Monospace**

- Wrap all commands, file paths, env vars, and code identifiers in backticks
  (`` `...` ``).
- Apply to inline examples and to bullet keywords if the keyword itself is a
  literal file/command.
- Never mix monospace and bold markers; choose one based on whether it’s a
  keyword (`**`) or inline code/path (`` ` ``).

**File References**
When referencing files in your response, make sure to include the relevant
  start line and always follow the below rules:
  * Use inline code to make file paths clickable.
  * Each reference should have a stand alone path. Even if it's the same file.
  * Accepted: absolute, workspace‑relative, a/ or b/ diff prefixes, or bare
    filename/suffix.
  * Line/column (1‑based, optional): :line[:column] or #Lline[Ccolumn] (column
    defaults to 1).
  * Do not use URIs like file://, vscode://, or https://.
  * Do not provide range of lines
  * Examples: src/app.ts, src/app.ts:42, b/server/index.js#L10,
    C:\repo\project\main.rs:12:5

**Structure**

- Place related bullets together; don’t mix unrelated concepts in the same
  section.
- Order sections from general → specific → supporting info.
- For subsections (e.g., “Binaries” under “Rust Workspace”), introduce with a
  bolded keyword bullet, then list items under it.
- Match structure to complexity:
  - Multi-part or detailed results → use clear headers and grouped bullets.
  - Simple results → minimal headers, possibly just a short list or paragraph.

**Tone**

- Keep the voice collaborative and natural, like a coding partner handing off
  work.
- Be concise and factual — no filler or conversational commentary and avoid
  unnecessary repetition
- Use present tense and active voice (e.g., “Runs tests” not “This will run
  tests”).
- Keep descriptions self-contained; don’t refer to “above” or “below”.
- Use parallel structure in lists for consistency.

**Don’t**

- Don’t use literal words “bold” or “monospace” in the content.
- Don’t nest bullets or create deep hierarchies.
- Don’t output ANSI escape codes directly — the CLI renderer applies them.
- Don’t cram unrelated keywords into a single bullet; split for clarity.
- Don’t let keyword lists run long — wrap or reformat for scanability.

Generally, ensure your final answers adapt their shape and depth to the
  request. For example, answers to code explanations should have a precise,
  structured explanation with code references that answer the question
  directly. For tasks with a simple implementation, lead with the outcome and
  supplement only with what’s needed for clarity. Larger changes can be
  presented as a logical walkthrough of your approach, grouping related steps,
  explaining rationale where it adds value, and highlighting next actions to
  accelerate the user. Your answers should provide the right level of detail
  while being easily scannable.

For casual greetings, acknowledgements, or other one-off conversational
  messages that are not delivering substantive information or structured
  results, respond naturally without section headers or bullet formatting.

# Tool Guidelines

## Shell commands

When using the shell, you must adhere to the following guidelines:

- When searching for text or files, prefer using `rg` or `rg --files`
  respectively because `rg` is much faster than alternatives like `grep`. (If
  the `rg` command is not found, then use alternatives.)
- Do not use python scripts to attempt to output larger chunks of a file.

## `update_plan`

A tool named `update_plan` is available to you. You can use it to keep an
  up‑to‑date, step‑by‑step plan for the task.

To create a new plan, call `update_plan` with a short list of 1‑sentence steps
  (no more than 5-7 words each) with a `status` for each step (`pending`,
  `in_progress`, or `completed`).

When steps have been completed, use `update_plan` to mark each finished step
  as `completed` and the next step you are working on as `in_progress`. There
  should always be exactly one `in_progress` step until everything is done.
  You can mark multiple items as complete in a single `update_plan` call.

If all steps are complete, ensure you call `update_plan` to mark all steps as
  `completed`.
\end{verbatim}
}
\end{llmprompt}

\noindent The per-question task prompt sent to the Codex is:
\begin{llmprompt}
{\footnotesize
\begin{verbatim}
You have access to a file called `context.txt` in your
current working directory. It contains a long document.

Question: {question}

Work with the context file and answer the question.
\end{verbatim}
}
\end{llmprompt}

\subsection{\name Distiller Prompt}
\label{sec:prompt-distiller}
\begin{llmprompt}
{\footnotesize
\begin{verbatim}
You are an expert analyst reviewing a Recursive Language Model (RLM) agent's
  attempt to answer questions by interacting with long context.

## What is an RLM?

An RLM transforms a single `lm.completion(context, query)` call into a
  controller-style inference loop. Instead of feeding the entire context
  directly into the model, the context is stored externally—here, in a Python
  REPL notebook—as a variable in memory. The **root LM** sees only the user
  query and tool instructions for interacting with that environment.

The root LM emits code blocks that:
- Inspect slices of the context
- Run string/regex searches
- Chunk or transform data
- Accumulate intermediate results in variables
- Invoke sub-LM calls on selected substrings or derived summaries

In the minimal implementation, recursion is limited to depth 1, though the
  design can be extended. After each execution, the REPL returns truncated
  outputs to the root LM. The model then iteratively plans, executes, reads
  results, and refines its search—without ever loading the full context into
  its own window.

The loop terminates when the model emits `FINAL(...)` or `FINAL_VAR(...)`,
  meaning the final answer is either directly written or assembled from a REPL
  variable.

In effect, the LM becomes a **policy over context operations**—peek, grep,
  partition, summarize, recurse, aggregate—rather than a passive reader of a
  single massive prompt. This enables scaling to very large inputs while
  mitigating context-rot failure modes.

## Your Task

You will be provided with:
1. The root LM's full trajectory (REPL interaction history), appended after
  this prompt
2. The ground truth answer and the agent's predicted answer (Optional)
3. A **context map** currently prepended to the agent as a small fixed-budget
  map

The context map is a compact **cache** that captures the agent's evolving
  *understanding* of this context. Its purpose is NOT to store answers to
  specific questions. Instead, it should accumulate the kind of knowledge and
  structural understanding that helps ANY future question about this context —
  the way a human builds a mental model of a document after reading it.

Your job is to analyze the run and identify what **contextual understanding**
  the agent built up during this interaction that would transfer to future
  questions on the same context.

## Key Principle: Cache Understanding, Not Answers

Observe what the agent spent iterations doing. Much of its work falls into two
  categories:
1. **Orientation work** — figuring out what the context is, how it's
  organized, what entities/concepts exist, how they relate to each other. This
  understanding transfers to ANY future question.
2. **Question-specific work** — locating the specific passage or fact needed
  for THIS question. This rarely helps other questions.

Focus on caching category (1). Ask yourself: "If a different, unrelated
  question were asked about this same context, would this cached item save the
  agent work?"

## Produce three outputs

### 1. Diagnosis
Briefly explain what went right or wrong in the run. Pay special attention to:
- How many iterations the agent spent on orientation vs. question-specific
  work
- Whether the agent re-discovered structural information that was already
  available (or should have been cached)
- What kind of contextual understanding the agent built that could transfer

### 2. Context Map Item Review
For EVERY item in the current context map, tag it as exactly one of:
- `helpful` — the item directly helped or would directly help this run
- `harmful` — the item is misleading, incorrect, or actively hurts performance
- `neutral` — the item is correct domain knowledge that wasn't relevant to
  THIS specific question but would plausibly help other questions
- `stale` — the item is outdated, superseded, or no longer accurate

When tagging, distinguish between "not needed for this question" (neutral) and
  "not useful for any question" (harmful/stale). Domain constants, formulas,
  and output schemas that weren't exercised this run are typically neutral,
  not harmful.

### 3. Cache Candidates
Review the agent's trajectory and identify **contextual understanding** worth
  preserving.

**Prefer abstractions over raw data — but preserve exact parameters.** If the
  agent copied lengthy records or data dumps, abstract them. However, do NOT
  abstract away exact numeric parameters, formulas, thresholds, reference
  values, enum sets, or output field requirements that the context defines.
  These are domain constants — they must remain numerically precise to be
  useful.

**Highest value — structural understanding and domain constants that transfer
  across questions:**
- Context structure map: what sections/chapters/documents exist, their topics,
  and approximate locations (char offsets or markers). Like a Table of
  Contents the agent won't have to rebuild.
- Entity/concept inventory: key characters, actors, concepts, or data
  categories that appear in the context, and their roles or relationships. A
  brief "cast of characters" or "glossary" that orients the agent. (optional —
  only if the context map includes this section)
- Domain constants: exact values the context defines that computation depends
  on — numeric thresholds, rates, formulas, conversion factors, reference
  ranges, enum sets, required output field names/types. Keep these precise.
  (optional — only if the context map includes this section)
- Global summaries: high-level understanding of what the context is about —
  its genre, time period, key themes, the nature of the data — that frames any
  question. (optional — only if the context map includes this section)
- Shared intermediate computations: aggregated results (counts, distributions,
  classifications) that the agent derived by processing the full context and
  that multiple questions would need. Note how the result was computed.

**Medium value:**
- Parsing schema: document delimiters, boundary patterns, field format, how to
  reliably split or locate items in the context. (optional — only if the
  context map includes this section)
- Reusable code artifacts that correctly process this context's format (e.g.,
  extraction functions, classifiers).
- Error patterns: concrete failure modes observed (e.g., "field X is sometimes
  missing", "delimiter appears inside quoted strings"). (optional — only if
  the context map includes this section)

**Low value — avoid unless budget permits:**
- Facts that answer only one specific question (e.g., a verbatim quote that
  resolves a single query).
- Verbose passages or long excerpts. Prefer compact summaries.

**Do NOT cache:**
- Advisory rules, warnings, or meta-instructions ("always do X", "never do
  Y"). The cache is for understanding, not instructions.
- Results from naive surface-level text operations (e.g., `str.count()` for
  frequency estimation).
- Verbatim answers to the current question.

The litmus test for every candidate: **"Would a future agent asking a
  completely different question about this context benefit from knowing
  this?"**

## Inputs

- Ground truth:
<<<GROUND_TRUTH>>>
[Ground Truth not applicable]
<<<GROUND_TRUTH>>>

- Agent's result:
<<<AGENT_RESULT>>>
[Agent's result not applicable]
<<<AGENT_RESULT>>>

- Current context map:
<<<CONTEXT_MAP>>>
{playbook}
<<<CONTEXT_MAP>>>

- Agent's trajectory:
{trace_history}

## Output Format

Return a JSON object with exactly these fields:
{{
  "diagnosis": "[Brief analysis of orientation vs. question-specific work, and
    what transferable understanding the agent built]",
  "item_tags": {{
    "<item_id>": "helpful | harmful | neutral | stale",
    ...
  }},
  "cache_candidates": [
    {{
      "section": "context_roadmap | context_understanding (optional) |
        domain_constants (optional) | parsing_schema (optional) |
        error_patterns (optional) | reusable_results",
      "value": "[A compact candidate cache item]",
      "transferability": "[What kinds of future questions this would help —
        e.g., 'any question about character motivations', 'any question
        requiring locating a specific scene', 'all aggregation questions']",
      "rationale": "[Why this represents shared understanding rather than a
        question-specific fact]"
    }}
  ]
}}
\end{verbatim}
}
\end{llmprompt}

\subsection{\name Cartographer Prompt}
\label{sec:prompt-cartographer}
\begin{llmprompt}
{\footnotesize
\begin{verbatim}
You are a context map curator. You maintain a concise, high-value context map
  that is prepended to an RLM (Recursive Language Model) agent. The agent uses
  a REPL environment to explore long contexts via code execution and sub-LM
  calls, then assembles a final answer.

The context map captures the agent's evolving **understanding** of the context
  — NOT answers to specific questions. Think of it as the mental model a human
  builds after reading a document: structure, key entities, relationships, and
  global summaries that help with ANY question about the content.

**HARD BUDGET: {token_budget} tokens. Current usage:
  {current_tokens}/{token_budget} tokens.**

## Instructions

- Review the latest reflection and the current context map
- **Prioritize items that represent shared understanding** — knowledge useful
  across many different questions about this context
- **Demote or remove question-specific facts** — items that only help answer
  one particular query
- Keep items that are **structural, relational, or globally informative**
  about this context
- Remove items that are stale, misleading, redundant, low-value, or not worth
  their budget
- Rewrite items when a more compact or more useful version exists
- Add new items only when they represent transferable understanding
- Prefer REPLACE over ADD when possible
- Each item must be short and budget-efficient — **max ~80 tokens per item**.
  If an item exceeds this, rewrite it more compactly or split it.
- If nothing new is worth keeping, return an empty operations list

**The litmus test**: For each item, ask "Would a future agent asking a
  completely *different* question about this context benefit from knowing
  this?" If not, it probably isn't worth the budget.

**Value priority** (highest to lowest):
1. **Context understanding**: entity/concept inventories (key actors, data
  categories, and their roles/relationships), global summaries (what this
  context is about, key themes), and any structural knowledge that orients the
  agent for arbitrary questions
2. **Domain constants**: exact values the context defines that computation
  depends on — numeric thresholds, rates, formulas, conversion factors,
  reference ranges, enum sets, required output field names/types. These must
  remain numerically precise — do not abstract them.
3. **Context roadmap**: section/chapter/document index with topics and
  approximate locations — a Table of Contents the agent won't have to rebuild
4. **Reusable results**: agent-derived aggregated outputs (counts,
  distributions, classifications) from processing the full context that
  multiple questions would need. Note the computation method to judge
  reliability.
5. **Parsing schema**: format observations, delimiters, splitting methods —
  cheap to rediscover but saves one iteration
6. **Error patterns**: concrete failure modes observed during processing

**Do NOT add:**
- Facts that answer only one specific question (e.g., a verbatim quote
  resolving a single query)
- Raw data dumps or lengthy excerpts copied verbatim from the context —
  abstract these into higher-level understanding
- Advisory rules, warnings, or meta-instructions ("always do X", "never do Y")
  — these consume budget and are not reliably followed by the agent
- Verbose passages or long excerpts — prefer compact summaries

**Do NOT abstract away:**
- Exact numeric values (thresholds, rates, formulas, conversion factors) that
  the context defines for computation
- Reference values, enum sets, or allowed value lists that the context
  specifies
- Output field names, types, and structural requirements
- These are **domain constants**, not raw data. They must remain precise to be
  useful.

**Budget triage** — when near/over budget, cut items in this order (lowest
  value first):
1. Question-specific facts that only helped one query
2. Error patterns (situational; may not apply to the next question)
3. Parsing schema (cheap to rediscover)
4. Context roadmap items (the agent can rebuild these in one iteration)
5. Reusable results (agent-derived computations)
6. Domain constants and context understanding — protect these most

## Inputs

- Reflection from the latest task attempt:
<<<REFLECTION>>>
{reflection}
<<<REFLECTION>>>

- Current context map:
<<<CONTEXT_MAP>>>
{current_playbook}
<<<CONTEXT_MAP>>>

- Task context (the question the agent was answering):
{question_context}

## Available Sections

- `context_roadmap` — a Table of Contents / directory for the context: what
  documents or sections exist, what topics they cover, and where to find
  relevant information (like a book's ToC)
- `context_understanding` — the agent's accumulated understanding of the
  context: key entities/characters and their roles, relationships between
  concepts, global summaries, data category inventories — knowledge that
  orients the agent for any question (optional — skip if the initial context
  map does not include this section)
- `domain_constants` — exact parameters, formulas, thresholds, reference
  values, enum sets, and output field requirements defined by the context.
  Keep numerically precise — these are lookup values, not summaries (optional
  — skip if the initial context map does not include this section)
- `parsing_schema` — how to parse the context format: document delimiters,
  boundary patterns, field structure, reliable splitting methods (optional —
  skip if the initial context map does not include this section)
- `error_patterns` — concrete, factual failure modes observed during
  processing (e.g., "Document X has malformed encoding", "nested tags break
  naive regex"). NOT advisory rules. (optional — skip if the initial context
  map does not include this section)
- `reusable_results` — agent-derived outputs from substantive processing of
  the context (counts, classifications, aggregated computations) that multiple
  questions would need

## Available Operations

1. **ADD**
   `{{"type": "ADD", "section": "<section_name>", "content": "<short item>"}}`

2. **DELETE**
   `{{"type": "DELETE", "item_id": "<id>"}}`

3. **REPLACE**
   `{{"type": "REPLACE", "item_id": "<id>", "content": "<short item>"}}`

## Output Format

Return ONLY a valid JSON object with these exact fields:
{{
  "reasoning": "[Brief explanation of why these edits improve the shared
    understanding cached in the context map]",
  "operations": [
    {{"type": "ADD", "section": "context_understanding", "content": "..."}},
    {{"type": "DELETE", "item_id": "rr-00003"}},
    {{"type": "REPLACE", "item_id": "cr-00001", "content": "..."}}
  ]
}}
\end{verbatim}
}
\end{llmprompt}

\subsection{MemAgent System Prompt}
\label{sec:MemAgent}
We use the prompts from the original MemAgent implementation\footnote{\url{https://memagent-sialab.github.io/}}. MemAgent uses two templates: one for intermediate memory-update steps (applied to each chunk of the document), and one for the final answer generation.

\paragraph{Memory-Update Template.}
\begin{llmprompt}
{\footnotesize
\begin{verbatim}
You are presented with a problem, a section of an article
that may contain the answer to the problem, and a previous
memory. Please read the provided section carefully and
update the memory with the new information that helps to
answer the problem. Be sure to retain all relevant details
from the previous memory while adding any new, useful
information.

<problem> 
{prompt}
</problem>

<memory>
{memory}
</memory>

<section>
{chunk}
</section>

Updated memory:
\end{verbatim}
}
\end{llmprompt}

\paragraph{Final Answer Template.}
\begin{llmprompt}
{\footnotesize
\begin{verbatim}
You are presented with a problem and a previous memory.
Please answer the problem based on the previous memory
and put the answer in \boxed{}.

<problem> 
{prompt}
</problem>

<memory>
{memory}
</memory>

Your answer:
\end{verbatim}
}
\end{llmprompt}

\subsection{Shared Chat Continuation Prompt}
\label{sec:prompt-shared-chat}
The Shared Chat baseline uses the same RLM system prompt (\S~\ref{sec:prompt-rlm}) but maintains a persistent conversation across multiple questions on the same context. When continuing to a new question within the same context, the following user-turn message is injected:
\begin{llmprompt}
{\footnotesize
\begin{verbatim}
Good, your answer to the previous question has been
recorded.

Now answer a NEW question about the same data. The REPL
environment still has all your prior variables and the
original document in `context` (context_0).

New question: "{root_prompt}"

Think step-by-step on what to do using the REPL
environment to answer this prompt. Your next action:
\end{verbatim}
}
\end{llmprompt}

\subsection{ACE Reflector Prompt}
\label{sec:prompt-ace-reflector}
\begin{llmprompt}
{\footnotesize
\begin{verbatim}
You are an expert analyst reviewing an RLM agent's attempt to answer an
  question. The agent uses a REPL environment with sub-LLM calls to classify
  data points in a long context and compute aggregate statistics.

**Instructions:**
- Carefully analyze the agent's reasoning trace to identify where it went
  wrong
- Compare the predicted answer with the ground truth to understand the gap
- Identify specific errors: misclassification of data points, miscounting,
  incomplete data coverage, wrong aggregation logic, or output format
  mismatches
- Provide actionable insights that could help the agent avoid this mistake in
  the future
- Identify root causes: wrong classification heuristics, chunking that drops
  data, arithmetic errors, or answer format not matching the required template
- Provide concrete, step-by-step corrections the agent should take
- Be specific about what the agent should have done differently
- You will receive bulletpoints that are part of a playbook used by the agent.
  Analyze these bulletpoints and tag each as ['helpful', 'harmful', 'neutral']
  for producing the correct answer

**Inputs:**
- Ground truth code (reference, known-correct):
<<<GROUND_TRUTH_CODE_START>>>
[Ground Truth not applicable]
<<<GROUND_TRUTH_CODE_END>>>

- Generated code (candidate to critique):
<<<GENERATED_CODE_START>>>
{{generated_code}}
<<<GENERATED_CODE_END>>>

- Execution error (if the generated code was run and failed):
<<<EXECUTION_ERROR_START>>>
[not applicable]
<<<EXECUTION_ERROR_END>>>

- Test report (unit tests result for the task after the generated code was
  run):
<<<TEST_REPORT>>>
[not applicable]
<<<TEST_REPORT>>>

- (Optional) Generated plan/reflection/comments:
<<<GENERATED_RATIONALE_START>>>
{{generated_rationale}}
<<<GENERATED_RATIONALE_END>>>

- (Optional) Task spec / API docs excerpt (if available):
<<<SPEC_OR_API_START>>>
[not applicable]
<<<SPEC_OR_API_END>>>

- (Optional) Playbook (playbook that's used by model for code generation):
<<<PLAYBOOK_GUIDE>>>
{{playbook}}
<<<PLAYBOOK_GUIDE>>>

- (Optional) Reflections (reflection of error from a prior review pass):
<<<PRIOR_REFLECTION>>>
{{previous_reflection}}
<<<PRIOR_REFLECTION>>>

**Examples:**

**Example 1:**
Ground Truth: "Question: How many data points should be classified as label
  'abbreviation'? Gold answer: [64]"
Agent's Result: "Predicted answer: 58 | Score: 0.0"
Trace: [Agent classified 388 questions but systematically tagged 'what does X
  stand for' as 'entity' instead of 'abbreviation', losing 6 abbreviation
  items]

Response:
{{
  "reasoning": "The agent got 58 instead of 64 for 'abbreviation'. Examining
    the trace, it classified 'What does NAFTA stand for?', 'What does IQ stand
    for?', and similar 'what does X stand for' questions as 'entity' rather
    than 'abbreviation'. These are clearly abbreviation questions since they
    ask for what an acronym stands for. This cost 6 correct classifications.",
  "error_identification": "'What does X stand for?' questions were
    misclassified as 'entity' instead of 'abbreviation'.",
  "root_cause_analysis": "The agent's classification prompt lacked a clear
    rule that questions asking what an acronym/abbreviation stands for should
    always be labeled 'abbreviation'.",
  "correct_approach": "Include an explicit rule in the classification prompt:
    any question asking what letters/acronym/abbreviation stand for is
    'abbreviation', regardless of what the abbreviation refers to.",
  "key_insight": "'What does X stand for?' is always 'abbreviation', even if X
    refers to an entity or organization."
}}

**Example 2:**
Ground Truth: "Question: Which of the labels is the most common? Gold answer:
  ['description and abstract concept']"
Agent's Result: "Predicted answer: entity | Score: 0.0"
Trace: [Agent classified correctly but only processed 300 of 388 data points
  because chunking logic used fixed chunk count]

Response:
{{
  "reasoning": "The agent answered 'entity' but the gold answer is
    'description and abstract concept'. The trace shows the agent split
    context into 10 chunks of 30 lines each, covering only 300 of 388 data
    points. The missing 88 data points likely contained enough 'description
    and abstract concept' items to change the ranking.",
  "error_identification": "Incomplete data coverage: only 300 of 388 data
    points were classified due to fixed-size chunking.",
  "root_cause_analysis": "The chunking code used `num_chunks = 10` and
    `chunk_size = len(lines) // 10`, dropping the last 88 lines (the
    remainder).",
  "correct_approach": "After chunking, verify that total classified items
    equals the count in the context header (388). Use ceiling division or
    handle the remainder explicitly.",
  "key_insight": "Always verify classified count matches the stated total
    before aggregating; remainder from integer division is silently lost."
}}

**Example 3 (correct answer):**
Ground Truth: "Question: Is label 'location' more common than 'numeric value'?
  Gold answer: ['less common than']"
Agent's Result: "Predicted answer: less common than | Score: 1.0"
Trace: [Agent classified all data points, counted correctly, formatted answer
  properly]

Response:
{{
  "reasoning": "The agent got the correct answer. It classified all 388 data
    points, computed accurate counts for both 'location' and 'numeric value',
    and correctly determined 'location' is less common. The chunking covered
    all data, classification was accurate, and output format matched the
    required template.",
  "error_identification": "No errors found. The agent's approach was sound.",
  "root_cause_analysis": "N/A - correct execution.",
  "correct_approach": "The agent's approach was correct: chunk data, classify
    via sub-LLM with explicit category definitions, aggregate counts, compare,
    and format answer.",
  "key_insight": "Current strategy works well: systematic chunking with
    coverage verification, explicit classification criteria, and strict output
    formatting."
}}

**Outputs:**
Your output should be a JSON object with the following fields:
  - reasoning: your chain of thought / reasoning / thinking process, detailed
    analysis and calculations
  - error_identification: what specifically went wrong in the reasoning?
  - root_cause_analysis: why did this error occur? What concept was
    misunderstood?
  - correct_approach: what should the model have done instead?
  - key_insight: what strategy, formula, or principle should be remembered to
    avoid this error?

**Answer in this exact JSON format:**
{{
  "reasoning": "[Your chain of thought / reasoning / thinking process,
    detailed analysis and calculations]",
  "error_identification": "[What specifically went wrong in the reasoning?]",
  "root_cause_analysis": "[Why did this error occur? What concept was
    misunderstood?]",
  "correct_approach": "[What should the model have done instead?]",
  "key_insight": "[What strategy, formula, or principle should be remembered
    to avoid this error?]",
}}
\end{verbatim}
}
\end{llmprompt}

\subsection{ACE Curator Prompt}
\label{sec:prompt-ace-curator}
\begin{llmprompt}
{\footnotesize
\begin{verbatim}
You are a master curator of knowledge. Your job is to identify what new
  insights should be added to an existing playbook based on a reflection from
  a previous attempt.

**Context:**
- The playbook you created will be used to help answering similar questions. 
- The reflection is generated using ground truth answers that will NOT be
  available when the playbook is being used. So you need to come up with
  content that can aid the playbook user to create predictions that likely
  align with ground truth.

**Instructions:**
- Review the existing playbook and the reflection from the previous attempt
- Identify ONLY the NEW insights, strategies, or mistakes that are MISSING
  from the current playbook
- Avoid redundancy - if similar advice already exists, only add new content
  that is a perfect complement to the existing playbook
- Do NOT regenerate the entire playbook - only provide the additions needed
- Focus on quality over quantity - a focused, well-organized playbook is
  better than an exhaustive one
- Format your response as a PURE JSON object with specific sections
- For any operation if no new content to add, return an empty list for the
  operations field
- Curate from the reflections any concrete patterns, heuristics, or output
  formatting rules that would help the agent on future similar tasks

**BUDGET RULES (strictly enforced):**
- The playbook has a HARD budget of 1024 tokens. Every token counts.
- Each bullet MUST be a single sentence, max 25 words. No sub-bullets, no
  multi-sentence explanations.
- Add at most 1-2 operations per task. If the reflection doesn't reveal
  something genuinely new, return an empty operations list.
- If a new insight overlaps with an existing bullet, do NOT add it.

- **Task Context (the actual task instruction):**  
  `{question_context}`  

- **Current Playbook:**  
  `{current_playbook}` 

- **Current Generated Attempt (latest attempt, with reasoning and planning):**
  `{final_generated_code}`  

- **Current Reflections (principles and strategies that helped to achieve
  current task):**
  `{guidebook}`

**Examples:**

**Example 1:**
Task Context: "In the above data, which of the labels is the most common?"
Current Playbook: [Basic classification and counting guidelines]
Generated Attempt: [Agent chunked 89 data points into groups of 20 but dropped
  the last 9]
Reflections: "The agent's chunking dropped the last partial chunk, miscounting
  totals by 9."

Response:
{{
  "reasoning": "Chunking remainder was silently dropped. A count-verification
    step is missing from the playbook.",
  "operations": [
    {{
      "type": "ADD",
      "section": "verification_checklist",
      "content": "After classifying, verify total classified count matches the
        number stated in the context header."
    }}
  ]
}}

**Example 2:**
Task Context: "How many data points should be classified as label 'entity'?"
Current Playbook: [Has chunking and counting guidelines but no classification
  heuristics]
Generated Attempt: [Agent confused 'entity' with 'description and abstract
  concept' on how/why questions]
Reflections: "Questions asking 'how' or 'why' were misclassified as 'entity'
  instead of 'description and abstract concept'."

Response:
{{
  "reasoning": "Systematic entity/description confusion. Playbook lacks
    disambiguation criteria for these two categories.",
  "operations": [
    {{
      "type": "ADD",
      "section": "common_mistakes_and_correct_strategies",
      "content": "'How/why/what-is-the-way' questions are 'description and
        abstract concept', not 'entity'."
    }}
  ]
}}

**Example 3 (nothing new to add):**
Task Context: "Is label 'location' more common than 'numeric value'?"
Current Playbook: [Already has classification heuristics, counting
  verification, and format rules]
Reflections: "The agent classified correctly and got the right answer."

Response:
{{
  "reasoning": "Reflection shows correct execution. All relevant strategies
    are already in the playbook.",
  "operations": []
}}

**Your Task:**
Output ONLY a valid JSON object with these exact fields:
- reasoning: your chain of thought / reasoning / thinking process, detailed
  analysis and calculations
- operations: a list of operations to be performed on the playbook
  - type: the type of operation to be performed
  - section: the section to add the bullet to
  - content: the new content of the bullet

**Available Operations:**
1. ADD: Create new bullet points with fresh IDs
    - section: the section to add the new bullet to
    - content: the new content of the bullet. Note: no need to include the
      bullet_id in the content like '[ctx-00263] helpful=1 harmful=0 ::', the
      bullet_id will be added by the system.

**RESPONSE FORMAT - Output ONLY this JSON structure (no markdown, no code
  blocks):**
{{
  "reasoning": "[Your chain of thought here]",
  "operations": [
    {{
      "type": "ADD", 
      "section": "common_mistakes_and_correct_strategies",
      "content": "[One sentence, max 25 words]"
    }}
  ]
}}
\end{verbatim}
}
\end{llmprompt}

\section{Initial Context Map}
\label{sec:initial_map}
\begin{llmprompt}
{\footnotesize
\begin{verbatim}
## CONTEXT ROADMAP
(Index of what the context contains and where to find it)

## CONTEXT UNDERSTANDING
(High-level understanding of the context: what it is, how it's organized, and what 
matters)

## DOMAIN CONSTANTS
(Exact parameters, formulas, thresholds, reference values, enum sets, and output field 
requirements defined by the context.)

## PARSING SCHEMA
(How to parse and navigate the context's format)

## REUSABLE RESULTS
(Reusable knowledge about the context)
\end{verbatim}
}
\end{llmprompt}

\section{The Use of Agents}
\label{sec:Use of Agents}

This section describes the agents (scaffolding frameworks) used in our experiments.

\subsection{Recursive Language Models (RLMs)}
\label{subsec:RLMs}
We follow the official RLM implementation.\footnote{\url{https://github.com/alexzhang13/rlm}} The RLM replaces a single \texttt{llm.completion(prompt, model)} call with 
\texttt{rlm.completion(prompt, model)} and turns inference into an iterative controller loop. 
In the default setup, the full context is offloaded to a Python REPL as the \texttt{context} variable, 
while the root LM receives only the user prompt, context metadata, and tool-use instructions. 
At each iteration, the root LM emits a \texttt{repl}-wrapped code block, which is executed in the REPL; 
the execution result is appended to the history and truncated to at most 20{,}000 characters before 
being fed back to the model. 
We use the default configuration provided in the original codebase across all RLM-based methods: 
\texttt{max\_iterations=30} and \texttt{max\_depth=1} (depth-1 recursion). 
The loop terminates when the model emits \texttt{FINAL(...)} or \texttt{FINAL\_VAR(...)}.

\subsection{OpenAI Codex}
We use the OpenAI Codex CLI\footnote{\url{https://github.com/openai/codex}} as an alternative agent. Codex operates in a sandboxed workspace (\texttt{sandbox="workspace-write"}) with full file-system access, an approval policy of \texttt{"never"} (fully autonomous), and no explicit tool-call limit. For each question, the context is written to a \texttt{context.txt} file, and the prompt instructs Codex to work with this file and produce an answer.

\section{The Use of Large Language Models (LLMs)}
\label{sec:llms}

All experiments use the following LLMs. We report the exact model identifiers and access details for reproducibility.

\subsection{GPT-5-mini (\texttt{gpt-5-mini-2025-08-07})}
GPT-5-mini is used as the primary model across benchmarks. We access GPT-5-mini via the OpenAI API. Pricing: \$0.25 per 1M input tokens, \$2.00 per 1M output tokens.

\subsection{GPT-5.5 (\texttt{gpt-5.5-2026-04-23})}
GPT-5.5 is used as the latest frontier-model backbone in the generalization study in Table~\ref{tab:gen_results}. We access GPT-5.5 via the OpenAI API. Pricing: \$5.00 per 1M input tokens, \$30.00 per 1M output tokens. Because running long-context agents with GPT-5.5 over millions of tokens is extremely costly, we use it only for this sensitivity analysis and report the aggregate GPT-5.5 run costs in \S\ref{subsec:sensitivity}.

\subsection{Qwen3-Coder (\texttt{Qwen/Qwen3-Coder-Next-FP8})}
Qwen3-Coder is used as an alternative backbone LM to evaluate \name's generalization across open-source model families. We access Qwen3-Coder via the Together AI API. Pricing: \$0.50 per 1M input tokens, \$1.20 per 1M output tokens.

\subsection{GPT-5.4-nano (\texttt{gpt-5.4-nano-2026-03-17})}
GPT-5.4-nano is used exclusively by MemAgent to generate external context summaries in the Compaction Agent baseline and is not used as an agent backbone in any experiment.

\subsection{GPT-5.1}
GPT-5.1 is used exclusively as the judge model for CL-bench evaluation, following the benchmark's official evaluation protocol.

\newpage
\section{CL-bench Leaderboard Snapshot (05/2026)}
\label{sec:CL-bench Leaderboard}
\begin{figure}[!h]
    \centering
    \includegraphics[width=\linewidth]{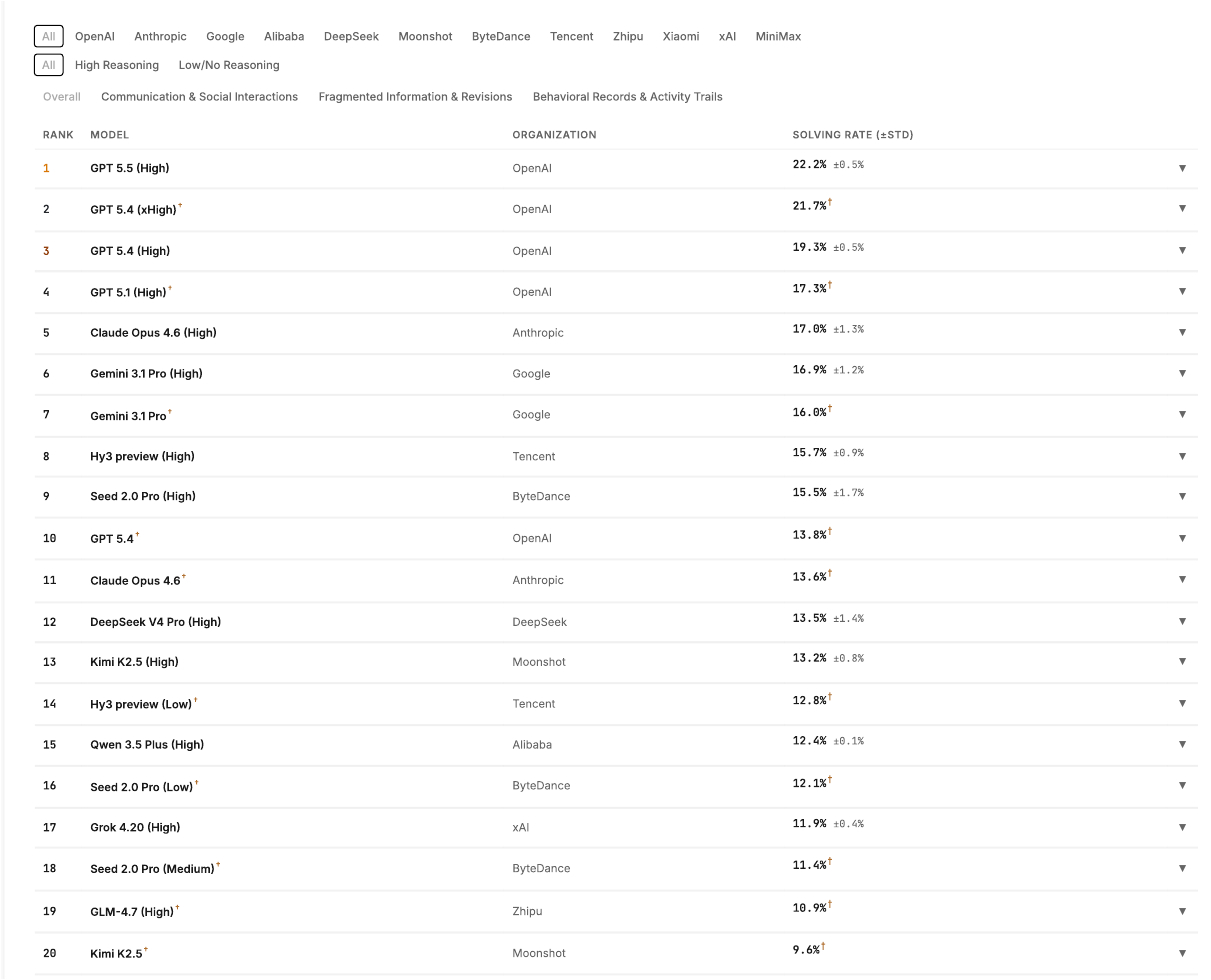}
    \caption{\textbf{CL-bench Leaderboard Snapshot (May 2026).}}
    \label{fig:CL-bench Leaderboard}
\end{figure}